\documentclass[10pt,twocolumn,letterpaper]{article}

\usepackage[preprint]{iccv}      %
\usepackage{appendix}
\usepackage{algorithm}
\usepackage{algorithmic}
\usepackage{enumitem}
\usepackage{makecell}

\definecolor{iccvblue}{rgb}{0.21,0.49,0.74}
\usepackage[pagebackref,breaklinks,colorlinks,allcolors=iccvblue]{hyperref}

\def\paperID{14624}
\def\confName{ICCV}
\def\confYear{2025}

\title{``\hspace{0.4mm}\textit{I Know It When I See It}\hspace{0.7mm}'':\\Mood Spaces for Connecting and Expressing Visual Concepts}

\author{
Huzheng Yang\textsuperscript{1} \quad
Katherine Xu\textsuperscript{1} \quad
Michael D. Grossberg\textsuperscript{2} \quad
Yutong Bai\textsuperscript{3} \quad
Jianbo Shi\textsuperscript{1} \\
\textsuperscript{1}UPenn \quad
\textsuperscript{2}CUNY \quad
\textsuperscript{3}UC Berkeley \\
\url{https://huzeyann.github.io/mspace/}
}

\begin{document}

\twocolumn[{%
\maketitle
\vspace{-30 pt}
\begin{center}
    \centering
    \includegraphics[trim=0in 3.3in .4in 0in, clip,width=\textwidth]{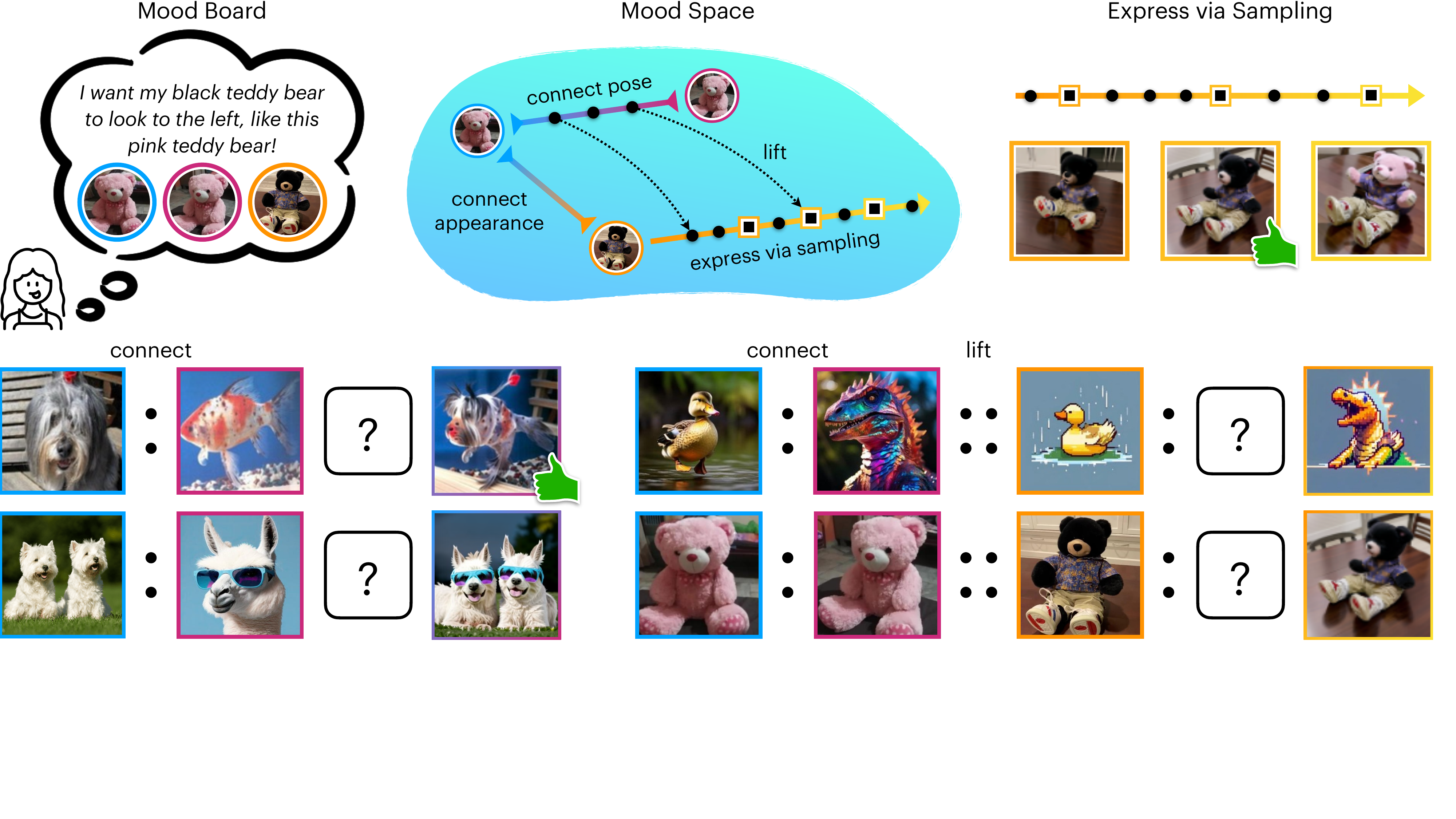}
    \vspace{4pt}
    \vspace{-20 pt}
    \captionof{figure}[Teaser]{Imagine a fictional animal that is a crossbreed of a sheepdog and goldfish.  No one has seen this animal, but we can recognize it `when I see it'.   We express such concepts via a Mood Board by curating exemplars that hint at our interest.  From a pre-trained feature (i.e., DINO), a latent Mood Space that is 50-100x smaller learns to squeeze out irrelevant features, find connections between the curated reference images, and decode back to any pre-trained feature space (i.e., CLIP).  The Mood Space is almost locally linear, thus supporting semantic image operations via simple vector algebra.  We demonstrate two Mood Space operations ``connect" (:) and ``lift" (::).  The ``connect" (:) operation makes a straight line in the Mood Space connecting two examples and decodes to a nonlinear curve in the image space.  It supports the image operation of object averaging.  The ``lift" (::) operation shifts the Mood Space curve by seeding it at a different reference sample.  It supports visual analogy and pose transfer. Multiple samples along the Mood Space curve provides diverse outputs tailoring to our individual `I know it when I see it' preferences.}
    \label{fig:teaser}
    \vspace{-5pt}
\end{center}%
}]

\begin{abstract}

Expressing complex concepts is easy when they can be labeled or quantified, but many ideas are hard to define yet instantly recognizable.  We propose a Mood Board, where users convey abstract concepts with examples that hint at the intended direction of attribute changes.   We compute an underlying Mood Space that 1) factors out irrelevant features and 2) finds the connections between images, thus bringing relevant concepts closer. We invent a fibration computation to compress/decompress pre-trained features into/from a compact space, 50-100x smaller. The main innovation is learning to mimic the pairwise affinity relationship of the image tokens across exemplars.   To focus on the coarse-to-fine hierarchical structures in the Mood Space, we compute the top eigenvector structure from the affinity matrix and define a loss in the eigenvector space. The resulting Mood Space is locally linear and compact, allowing image-level operations, such as object averaging, visual analogy, and pose transfer, to be performed as a simple vector operation in Mood Space.  Our learning is efficient in computation without any fine-tuning, needs only a few (2-20) exemplars, and takes less than a minute to learn.

\end{abstract}
\vspace{-10pt}
    
\section{Introduction}
\label{sec:intro}
\vspace{-6pt}

How does one express a concept that one has in mind?
Expressing a concept is easy if it can be named, categorized in exact labels, quantifiable attributes, or described in words.  But for concepts that are harder to describe, they can only be ``recognized when I see them." What can we do?

We designate a collection of images from the user as a Mood Board to express concepts that break free of the narrow categorical way of quantifying data.   We draw inspiration from mood boards, widely used in the design community \cite{cassidy2011mood, mcdonagh2004mood, shi2025brickifyenablingexpressivedesign} to ensure everyone shares the same vision.

A Mood Space is a local latent space built from the Mood Board to capture the concept and a variation that the user wants to control.  There are two key technical properties.  First, a Mood Space is a tight and compact latent space, so dense sampling leads to a valid image representing the desired concept.  Second, a smooth path in the Mood Space should lead to a smooth transition in the sampled images.

The primary control is for a user to express the mood based on affinity, or who is close to who.  This affinity view of data says we don't care how the data is described (in attributes, labels); we only care about the affinity relationship to objects of interest that become more prominent based on a set of context images. Sampling the mood means looking for data that have similar affinity relationships.

Pre-trained feature spaces such as CLIP or DINO, have demonstrated excellent performance in capturing both semantic meaning, and part vs. whole structure. 
We learn a mapping from the DINO embedding space to an equivalent feature space called the Mood Space that is 50-100x smaller.
The objective is to find the smallest feature space that preserves the pairwise affinity relationships of image tokens within the context set.

The affinity matrix itself cannot distinguish relevant and irrelevant relationships. It is the top eigenvectors of the affinity matrix and their ordering that reveal the relevant and irrelevant parts of the relationship. Therefore, we define a loss on the eigenvectors.

We demonstrate our Mood Space by connecting distinct visual concepts, such as creating hybrid animals or fictional characters, as well as producing consistent outputs when traversing different paths in this Mood Space. Furthermore, we extend this idea to visual analogies by first connecting a pair of concepts, and then lifting that connection to another reference object. The main contributions of this work are:
\begin{enumerate}[noitemsep, nolistsep]
    \item We introduce the interface of a Mood Board to capture hard-to-describe visual concepts. We learn a Mood Space to connect concepts, and allow users to perform semantic operations on images like interpolation and visual analogy using simple vector algebra.
    \item We distinguish relevant and irrelevant features by analyzing the top eigenvectors of the affinity matrix among tokens across context images.
    \item We propose a compact Mood Space, 50-100x more compressed than a pretrained DINO or CLIP feature space.
    \item We showcase efficient learning with 4-layer token-wise MLPs that require only 2-20 example images and trains in under a minute.
\end{enumerate}

\section{Related Work}
\label{sec:related_work}

\begin{figure*}[!th]
 \vspace{-10 pt}
    \centering
    \includegraphics[trim=0in 7.8in 11.5in 0in, clip,width=\textwidth]{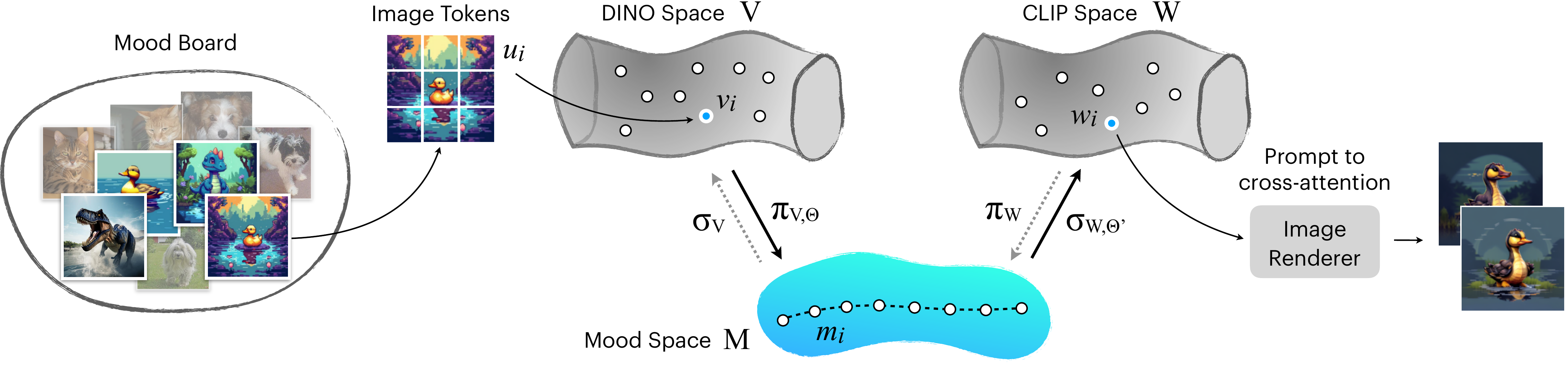}
    \vspace{-20 pt}
    \caption{We propose a context-specific DINO-to-CLIP mapping via a compressed Mood Space.   First, we compute per-patch image token embedding space, $V=\Bbb{R}^D$ using DINO.  We construct a low dimensional Mood Space $M$, with one map $\pi_V: V\to M$, and a second map $\sigma_W: M \to W$.  Intuitively, we think of $\pi_V: V \to M$ as a fiber bundle, or more generally, a fibration, that removes irrelevant feature spaces and brings closer the connection between the relevant tokens.   The base space $M$ parametrizes the variation we want to control.   The fibers, in other words, the pre-image $\pi_V^{-1}(m)$ of points $m\in M$, represent those aspects of the image tokens that we want to remain fixed. 
    Finally, we use a CLIP-conditioned image synthesis to render an output image.
    }
    \label{fig:method_pipeline}
    \vspace{-10 pt}
\end{figure*}

\subsection{Controllable Image Generation and Editing}

Recent advances in text-to-image diffusion models \cite{balaji2022ediff, betker2023improving, chen2023pixart, pernias2023wurstchen, ramesh2022hierarchical, rombach2022high, yu2022scaling, podell2023sdxl, sauer2025adversarial, luo2023latent, razzhigaev2023kandinsky} have significantly elevated the landscape of image generation. While such models enable the creation of highly realistic and diverse visual content, they sometimes struggle to synthesize images that follow user intentions from text prompts alone, leading to the development of methods for customizing and controlling the generation process. One such direction is building personalized diffusion models \cite{ruiz2023dreambooth, shi2024instantbooth, wang2024stableidentity, gal2022image, kumari2023multi, gal2023encoder, han2023svdiff, dong2022dreamartist}, which aims to generate image variations of a given concept or identity from one or few examples. Additional approaches \cite{cao2023masactrl, feng2024dit4edit, hertz2022prompt, tumanyan2023plug, brooks2023instructpix2pix} perform desired image edits by adjusting the self-attention or cross-attention from text instructions. Prior work \cite{kwon2022diffusion, park2023understanding, wu2023latent} has also explored finding semantic directions in the latent space of diffusion models \cite{chen2023pixart, rombach2022high, podell2023sdxl, sauer2025adversarial, esser2024scaling}, such as for the noise space \cite{samuel2023norm, brack2023sega}, weight space \cite{dravid2024interpreting, gandikota2024concept}, and text embedding space \cite{baumann2024continuous}.

There is another line of related work in image morphing \cite{wolberg1998image, aloraibi2023image, zope2017survey} that interpolates the latent space of generative models, such as GANs \cite{pan2021exploiting} and diffusion models \cite{zhang2024diffmorpher, he2024aid, wang2023interpolating, yang2023impus}. A key challenge is the unstructured image manifold learned in diffusion models, which may result in unsmooth transitions or inconsistent semantics during interpolation.

Perhaps most related to the present work is a user interface based on the manipulation of design tokens \cite{shi2025brickifyenablingexpressivedesign}. This work presents a user interface developed with extensive designer input that combines explicit visual prompts, as well as text and layout information. However, the goal is not to navigate the embedding space with samples but to address specific well-defined aspects. Each aspect is handled separately with a separate method such as extracting a color pallet from some image with statistics, or using ``imaginative tokens" to drive the layout. The goal is to create a specific design rather than navigate the embedding space.

\subsection{Manifold Discovery}

Mood Space is extracted by discovering a nonlinear subspace or manifold associated with the intended concept in the ambient feature space. This builds on a rich history of manifold discovery techniques, including t-SNE \cite{JMLR:v9:vandermaaten08a}, which emphasizes local relationships for visualization, and UMAP \cite{mcinnes2018umap}, which balances local and global structures for dimensionality reduction. Additionally, normalized cuts (Ncut) \cite{shi2000normalized} provide a graph-based approach to uncovering hierarchical structures through the eigenvectors of affinity matrices, where the most relevant feature is encoded in the top $k$ Ncut eigenvectors. The hierarchy of Ncut eigenvectors align with our use of mood space: we want to extract the most relevant concepts across the mood board images.

\section{Method}
\label{sec:method}

\begin{figure*}[!th]
 \vspace{-20 pt}
    \centering
    \includegraphics[trim=0in 8.1in 0.4in 0in, clip,width=\textwidth]{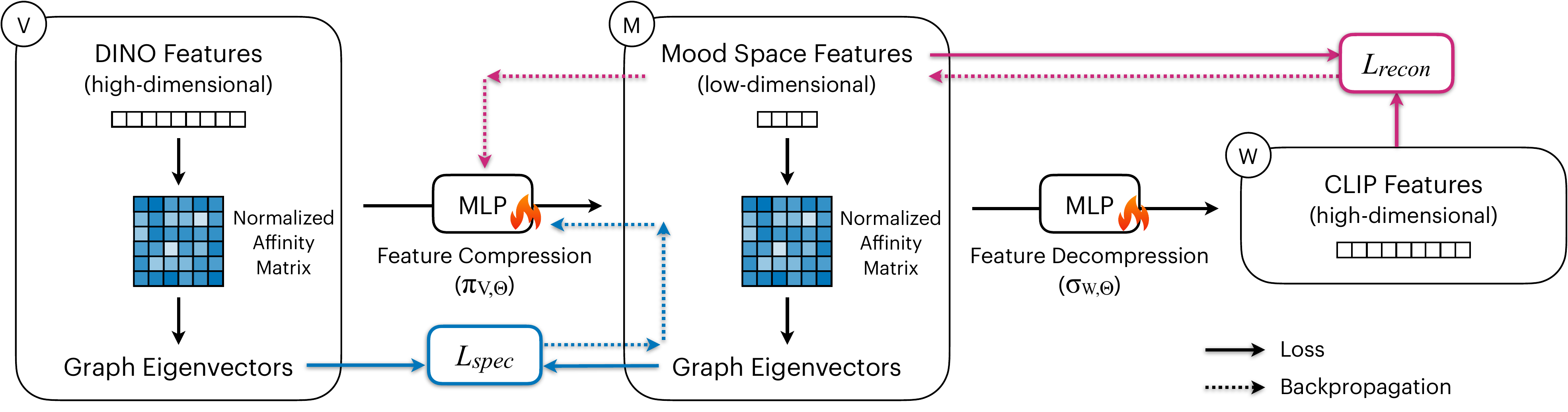}
    \vspace{-20 pt}
    \caption{ We collect image patch tokens across the context images and compute affinity relationships based on the pre-trained features in $V$ (DINO).  Our insight is that the token-pairwise affinity and its top eigenstructure contain part-whole hierarchical substructures that we want to preserve in $M$ (Mood Space).   A crucial observation is that the affinity relationship is representation agnostic, thus allowing us to define a loss function on affinity and backprop to train the representation in M, without feature alignment. Furthermore, by row-normalization of the affinity matrix and the top eigenvectors, we bring up the relevant concepts across images, making the connection easier to find. We train a token-wise MLP to decode from $M$ to $W$ (CLIP). }
    \label{fig:method_architecture}
    \vspace{-15 pt}
\end{figure*}

\subsection{Mood Space from Mood Board}

The human is asked to curate the Mood Board by specifying context images to express the visual concept. These images can come from a semantically similar domain (i.e., pets), where we want to construct a sequence of samples (crossbreed) connecting two exemplars (dog to fish).  Or, they could be two remotely related domains where we want to imagine how the examples can be connected visually.

\subsection{Effective Fiber Bundles}

\begin{figure}[!h]
    \centering
    \includegraphics[trim=0in 10.1in 10.55in 0in, clip,width=\columnwidth]{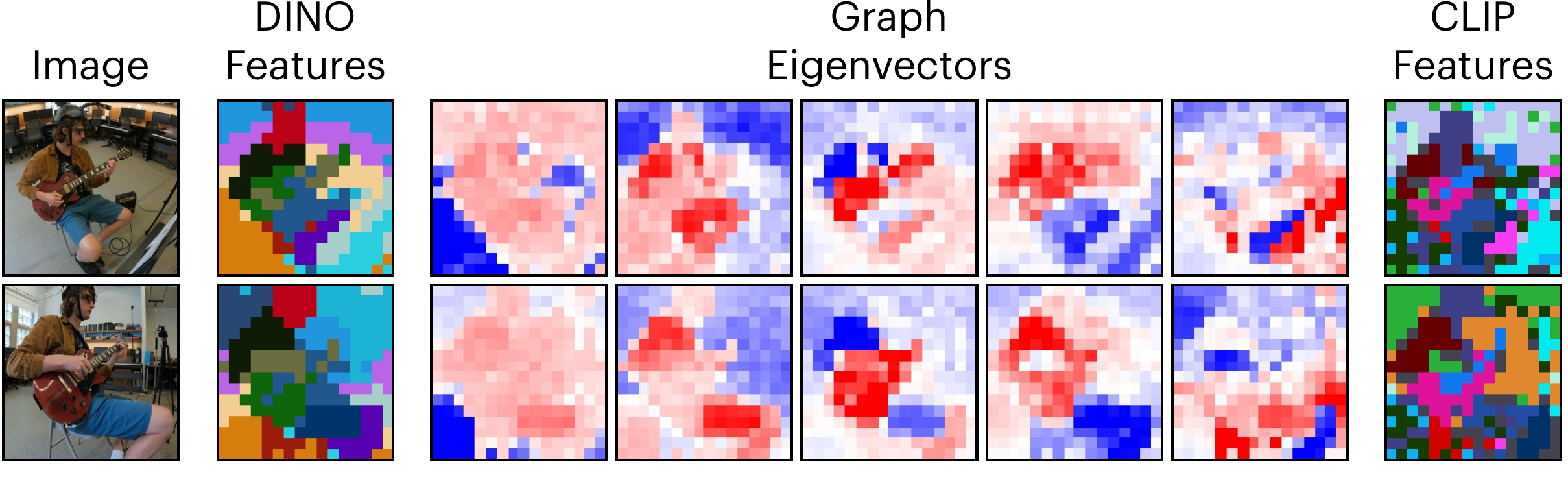}
    \vspace{-20 pt}
    \caption{For the two images shown left, we collect all the image and class tokens, compute their DINO features, and construct a row-normalized token affinity matrix  $S_V \in {\Bbb{R}}^{(2\times 256) \times (2\times 256)}$.  We compute the top 5 graph eigenvectors: $E(S_V)  \in {\Bbb{R}}^{ (2\times 256) \times 5} $.  Each column of $E(S_V)$ can be reshaped into images, and visualized as shown in the middle.   $E(S_V)$ are orthogonal to each other and encode hierarchical part-whole relationships, focusing on the relevant foreground objects.  Furthermore, we obtained part-level correspondence for free because we used DINO features to construct the affinity. We decode the compact space $M$ to the CLIP space $W$, shown on the right.  
    }
    \label{fig:method_feature_eigvec}
    \vspace{-15 pt}
\end{figure}

To construct the Mood Space shown in Figure \ref{fig:method_pipeline}, we start with a collection of context images $\mathcal{I} = \{ I_1, \ldots I_N\}$. Each image $I\in \mathcal{I}$ is broken into $16 \times 16=256$ patches $u$ of $16 \times 16$ pixels so that $u\in U=\Bbb{R}^{16\times16=256}$. Thus, we use $N\times 256$ image patches in $U$. We use the ViT tokenizer. 

We use two different pre-trained embeddings, DINO \cite{caron2021emerging} and CLIP \cite{radford2021learning}.
While the DINO embedding space captures semantic relationships and is better at correspondence, CLIP is better at representing semantics aligned with text. 

Suppose $V=\Bbb{R}^D$ is a $D$ dimensional DINO embedding space, and $W=\Bbb{R}^D$ is a $D$ dimensional CLIP embedding space. Let $T_V: U\to V$ map image tokens $U$ to DINO embeddings $V$, and similarly, $T_W: U\to W$ to CLIP embeddings. Let $v_i = T_V(u_i)$, and $w_i = T_W(w_i)$ for $1\leq i \leq 256\times N$. We want to construct a low dimensional Mood Space $M$ and two token level maps $\pi_V: V\to M$ and  $\sigma_W: M \to W$. Intuitively, we think of $\pi_V: V \to M$ as a fiber bundle, or more generally a fibration.

The base space $M$ parametrizes the variation we want to control. For example, if we have a collection of animal images with different backgrounds, we may want to focus on only the foreground animal's shape while disregarding the irrelevant background.
The fibers, in other words, the pre-image $\pi_V^{-1}(m)$ of points $m\in M$, represent those aspects of the image tokens that we want to remain fixed.
We do {\bf not} formally claim that $(V, M, \pi_V)$ is a fiber bundle (i.e., local product). We are simply using this as a guide; thus, we will refer to them as \emph{effective fiber bundles}.

\subsection{Learning Mood Space and Maps}

The Mood Space $M$ is a space that allows us to sample image tokens in a controlled way. We construct this Mood Space by considering the embedded tokens $v_i \in V, w_i \in W$ of our Mood Board. Because we want the Mood Space to be compact, we compute the effective dimension of the data in $V$ using an MLE estimation \cite{NIPS2004_74934548}. For a representative sample of our data, we found this dimension to be $G=6$ to $22$, depending on the Mood Board samples. We use $G$ as the target dimension of our Mood Space $M$. 

To find the Mood Space $M$ shown in Figure \ref{fig:method_architecture}, we consider a parametrized family of functions $\pi_{V,\Theta}$ implemented as a 4-layer MLP with parameters $\Theta$ and $G$ output units.
To compute $M$, we simply find an appropriate loss function to optimize over possible $\Theta$.  

We are trying to preserve the pairwise relationship between sample tokens $v_i$ based on their affinity matrix.  We compute the row normalized affinity matrix $S_V$ between all the tokens $v_i \in V$, with $(S_V)_{i,j}$ using RBF kernel with $\kappa e^{-||v_i-v_j||^2/h}$, followed by row normalization. For each $\Theta$, we can use $\pi_{V,\Theta}$ to project the sample points into $M$ and compute the affinity matrix $S_{M,\Theta}$. Since the affinity matrix itself cannot distinguish the relevant relationships across tokens that we want to connect vs. the accidental relationships among irrelevant features, we need a method to distinguish the relevant and irrelevant relationships.

\paragraph{Spectral Graph Embedding Loss.}

We use spectral graph embedding shown in Figure \ref{fig:method_feature_eigvec} to preserve a measure of graph eigenspace structure, which brings out relevant relationships in the top eigenvectors.
From the normalized affinity $S_V \in {\Bbb{R}}^{(N\times 256) \times (N\times 256)}$, we compute the top $k$ eigenvectors with the largest $k$ eigenvalues, and collect them into a matrix $E(S_V)  \in {\Bbb{R}}^{ (N\times 256) \times k} $.  The columns of $E(S_V)$ are vectors with values from -1 to 1, representing coarse-to-fine hierarchical partition.  Each row of $E(S_V)$ is a ${\Bbb{R}}^k$ vector that maps samples $v_i$ to its (graph) spectral embedding space.

There are two properties of this embedding that are crucial. First, the columns of $E(S_V)$ are orthogonal, allowing precise control of different aspects of the hierarchical structure we want to connect.
Second, the eigenvector values continuously vary, which is great for preserving and quantifying sample variations in the subspace we care about. Since we aim to find connections between samples within a context, these properties give us the precise tool to constrain the Mood Space.

Now that we have specified the primary structure in $V$, via $E(S_V)$, we can compute the same structure in $M$. That is, we can project the samples $\{v_i\}_{i=1}^{N\times 256}$ into $M=\Bbb{R}^G$ via the map $\pi_{V,\Theta}$ to get the image of the samples $\{m_{i,\Theta}\}_{i=1}^{N\times 256}$ dependent on the parameters $\Theta$.  Using these points we can compute the affinity matrix as above to obtain $S_{M,\Theta}$. Following that, we can again compute the spectral embedding via the map $E(M,\Theta)$. Note that unlike the spectrally embedded samples directly from $V$, these points depend on $\Theta$. We can now express our primary spectral loss term:

$$L_{\mbox{spec}}(\alpha,\Theta)= dist_{ev}(E(S_V), E(M, \Theta))$$
where $dist_{ev}$ is a vector similarity measure (\cref{alg:eigvec_loss}).

\begin{algorithm}[H]
\caption{Spectral Graph Embedding Loss}
\label{alg:eigvec_loss}
\begin{algorithmic}[1]
\REQUIRE $E(S_V)$ and $E(S_{M_{\Theta}}) \in \mathbb{R}^{(N \times 256)\times k}$

\textbf{for} {$i \in {4, 8, 16, 32, ..., k}$} \textbf{do}

\quad 1: Extract the top $i$ eigenvectors:

\qquad $E(S_V)_{:i} = E(S_V)[:, \text{1:}i]$

\qquad $E(S_{M_{\Theta}})_{:i} = E(S_{M_\Theta})[:, \text{1:}i]$

\quad 2: Compute the vector similarity loss:

\quad $\text{loss}_i = || E(S_V)_{:i} E(S_V)_{:i}^T - E(S_{M_\Theta})_{:i} E(S_{M_\Theta})_{:i}^T ||^2$

\textbf{end for}

\textbf{Return} \quad $L_{spec} = \sum_i \text{loss}_i$

\end{algorithmic}
\end{algorithm}

\paragraph{Additional Manifold Regularization.}
It would be desirable to construct the map $\pi_{V,\Theta}$ so that the Mood Space $M$ is locally linear.
If so, then at least for local neighborhoods, the geodesic paths between points become straight lines, and we can use the vector space structure to parameterize paths in $M$. As a result, we include a curvature loss term $L_{\mbox{curv}}(\Theta) = \Sigma_{i=0}^{N\times 256}||R(i,\Theta)||^2$ which estimates the Riemannian curvature tensor $R$ around each sample $m_{\Theta, i}$ in $M_{\Theta}$.  Another term ``repulsive force" prevents the data from bunching up on the edges of the embedding space, 
$L_{\mbox{rep}}(\Theta) = \Sigma_{i \neq j} 1/({||m_{i,\Theta} - m_{j,\Theta}||^2 + \epsilon})$. After projection down to $M$, recall the samples are lifted back to the CLIP embedding space $W$ via another 4-layer MLP $\sigma_{W,\Theta'}$ where $\Theta'$ is another set of parameter. Reconstructed CLIP embedding $\sigma_{W,\Theta'}(M_{\Theta})$ is then rendered out via a conditional generative model. As a result, we need a reconstruction loss $L_{\mbox{recon}}(\Theta, \Theta') = || W - \sigma_{W,\Theta'}(M_{\Theta}) ||^2$. Finally, we include a diversity term measured by the statistical variance of the samples $L_{\mbox{var}} ({\Theta}) = || \frac{1}{N\times256} M_\Theta^T M_\Theta - I ||^2$.

Putting it all together, the loss function is:
\[
\begin{aligned}
L(\Theta, \Theta') &= L_{\mbox{spec}}(\Theta) + \lambda_1 L_{\mbox{curv}}(\Theta) + \lambda_2 L_{\mbox{rep}}(\Theta) \\
&\quad + \lambda_3 L_{\mbox{recon}}(\Theta, \Theta') + \lambda_4 L_{\mbox{var}}(\Theta)
\end{aligned}
\]

\vspace{-4px}

\begin{algorithm}[h]
\caption{Learning Mood Space Mappings}
\label{alg:loss}
\begin{algorithmic}[1]
\REQUIRE Pre-trained embeddings: DINO and CLIP

\STATE \textbf{Feature Extraction:}

\quad $V \in \mathbb{R}^{(N \times 256) \times D_V}$: DINO embedding space

\quad $W \in \mathbb{R}^{(N \times 256) \times D_W}$: CLIP embedding space

\quad $G = \text{MLE}(V)$: estimate intrinsic dimension

\STATE \textbf{Mood Space and Mappings:}

\quad $M \in \mathbb{R}^{(N \times 256) \times G}$: Mood embedding space.

 Trainable mappings (point-wise MLP):

\quad $\pi_{V, \Theta}: V \to M$ and $\sigma_{W, \Theta'}: M \to W$.

\STATE \textbf{Affinity and Ncut:}

 Affinity matrix $S_V$ and $S_{M_{\Theta}} \in \mathbb{R}^{(N \times 256)\times(N \times 256)}$:
 
\quad $(S_V)_{i,j} = \kappa e^{-||v_i - v_j||^2/h}$

\quad $(S_{M_{\Theta}})_{i,j} = \kappa e^{-||m_{\Theta,i} - m_{\Theta,j}||^2/h}$

Compute Ncut \cite{shi2000normalized} top $k$ eigenvectors 

\quad $E(S_V)$ and $E(S_{M_{\Theta}}) \in \mathbb{R}^{(N \times 256)\times k}$

\STATE \textbf{Spectral Graph Embedding Loss:}

\quad \quad $L_{spec} = dist_{ev}(E(S_V), E(S_{M_{\Theta}}))$  \hfill (Algorithm 1)

\STATE \textbf{CLIP Embedding Reconstruction Loss:}

\quad \quad $ L_{recon} =|| W - \sigma_{W,\Theta'}(M_{\Theta}) ||^2$

\STATE \textbf{Regularizer on Mood Space:}

Riemannian curvature: $L_{\mbox{curv}} = \Sigma_{i=0}^{N\times 256}||R(i, \Theta)||^2$

Replusion: $L_{\mbox{rep}} = \Sigma_{i \neq j} 1/({||m_{\Theta,i} - m_{\Theta,j}||^2 + \epsilon})$

Zero covariance: $L_{\mbox{var}} = || \frac{1}{N\times256} M_\Theta^T M_\Theta - I ||^2$ 

\end{algorithmic}
\end{algorithm}

\subsection{Token Path Lifting}

For a pair of image tokens, $u_{A_1}$ and $u_{A_2}$ we compute a path connecting them, and generate CLIP embeddings for points along the path.  
We project these two image tokens into the Mood Space $M$ via the projection map $\pi_V$. Then, we connect these two projected points in Mood Space with a path $\gamma(t): [0,1]\to M$ from $m_{A_1}=\gamma(0)$ to $m_{A_2}=\gamma(1)$. Specifically, we take a naive path in $M$, $\gamma(t) = m_{A_1} + t*(m_{A_2}-m_{A_1})$. We say $\hat{\gamma}:[0,1]\to W$ is a lifting of $\gamma$ starting at $w_{A_1}$ if $\hat{\gamma}(0)=w_{A_1}$ and $\pi_W(\hat{\gamma}(t))=\gamma(t)$. In other words, $\gamma$ defines the path in Mood Space $M$, while $\hat{\gamma}$ defines the path in CLIP embedding space $W$.
We have constructed the section mapping $\sigma_W$ precisely for this purpose of lifting paths:
lift each point on the path in $M$ back to the CLIP space $W$ via $\sigma_w$: $\hat{\gamma}(t) = w_{A_1} + t*\sigma_w( m_{A_2}-m_{A_1})$. 

Suppose we change the starting point $w_{A_1}$ to a different but related token $w_{B_1}$. The difference between the two tokens could arise due to appearance, style, or shape. By lifting the curve in $M$ to a different starting point in $W$, we achieve a {\bf visual analogy} simply by: $\hat{\gamma}(t)^{B_1} = w_{B_1} + t*\sigma_w( m_{A_2}-m_{A_1} )  $.

\subsection{From Token Path to Image Path}

We want to promote the token path lifting to an image path lifting.
We will represent each image as a set of token clusters and find the cross-image correspondence between them using DINO features.
Then, we define a drift direction for each pair of corresponding token clusters in the Mood Space.  We move each token according to its drift direction to create an image path.
To achieve an image-level analogy for $A_1 : B_1 :: A_2 : B_2$, we first connect the path between $A_1$ and $B_1$. Then, we lift the token path from $A_1$ to $A_2$.

\subsection{Producing the Final Image}

Mood Spaces facilitate expressing and connecting visual concepts that a user has in mind. Specifically, our method produces CLIP image tokens that capture the essence of the Mood Board. This set of image tokens serves as inputs to a diffusion model conditioned on image prompts \cite{ye2023ip} to render the final image. Moreover, there are several ways in which a user can obtain context images for the Mood Board: gathering images from an existing dataset, finding images from the Internet, or using text-to-image generation.

\section{Experiments}
\label{sec:experiments}

We evaluate our method through two main experiments that demonstrate the key properties of our Mood Space: (1) its ability to find smooth paths connecting distinct visual concepts, and (2) its capability to perform consistent visual analogies through different paths. Our experiments show that the Mood Space provides better control and more meaningful interpolations compared to the baseline, which is linear interpolation in the CLIP embedding space $W$.

\subsection{Finding Paths to Connect Concepts}
\label{sec:eval_connect_concepts}

\textbf{Task.}
Visual concepts lie in different parts of a feature space. We want to show that we can find a smooth path connecting distinct visual concepts. We demonstrate finding good paths via interpolation between two input images. Smoothness means the interpolated images should change gradually between consecutive frames, blending the semantics from the two input images.

\begin{figure*}[!th]
    \centering
    \includegraphics[trim=0in 6.2in 8.55in 0in, clip,width=\textwidth]{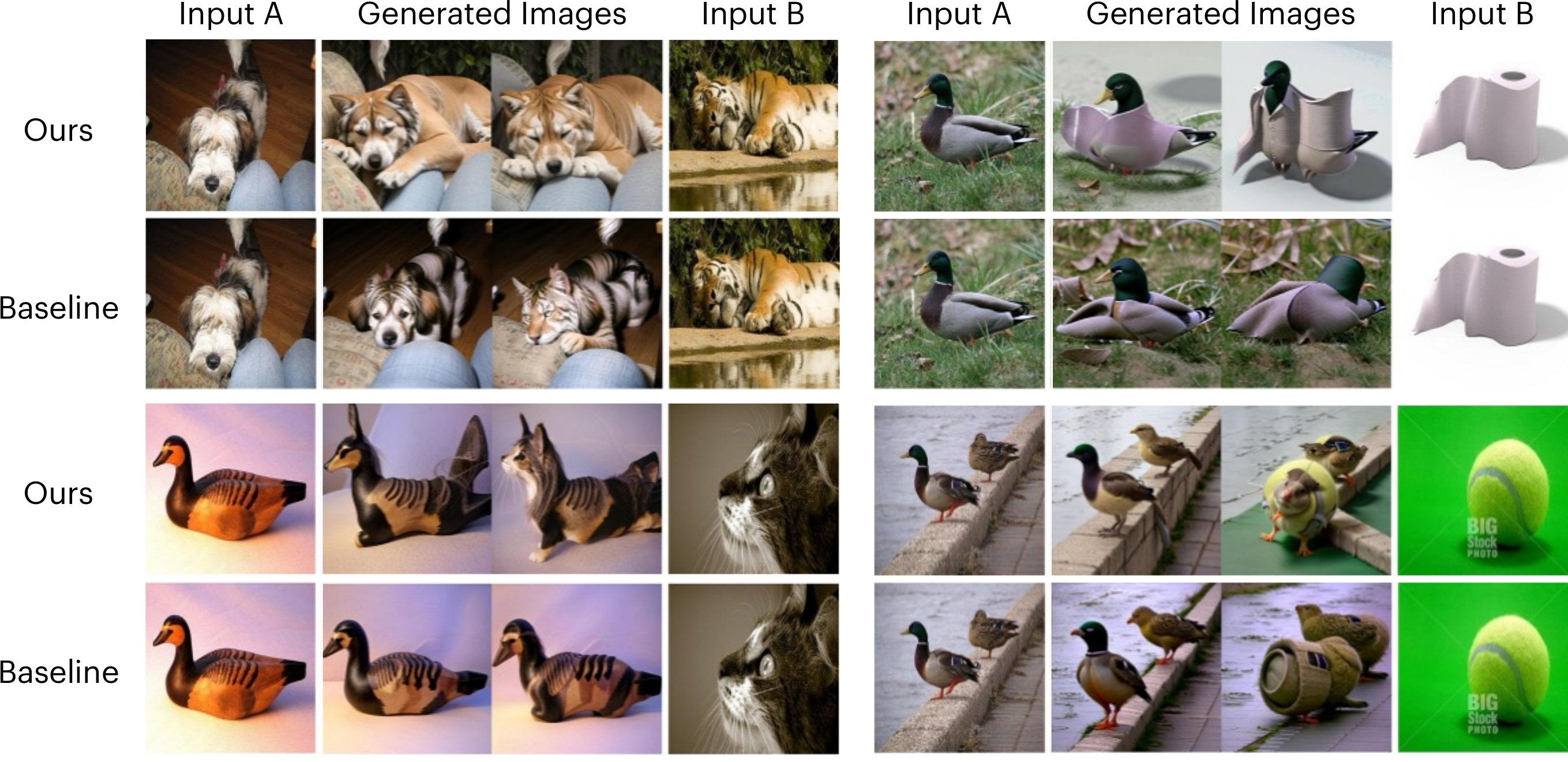}
    \vspace{-20 pt}
    \caption{Connecting concepts with inputs A and B. We compare our interpolation in Mood Space with the baseline linear interpolation in CLIP embedding space. We selected two samples using a interpolation weight $t$ of 0.5 and 0.6, intending to capture the hybrid object. Our method brings the two concepts closer, creating a hybrid version, instead of merely copying one object or the other. We succeed in both within-domain examples, like connecting a dog and a tiger, as well as different-domain examples, such as a bird and a paper roll.}
\label{fig:image_interpolation}
\end{figure*}

\noindent \textbf{Qualitative Results.}
Figure \ref{fig:image_interpolation} compares our method and the baseline on connecting distinct visual concepts. For the baseline, we use linear interpolation on CLIP image embeddings. We observe that the baseline often generates images similar to the first input image, while struggling to integrate visual details from the second input image. Our method connects the essence of both concepts and allows the user to express what they want by sampling multiple generations.

\noindent \textbf{Quantitative Results.}
We evaluate the smoothness of the path connecting two visual concepts using a dataset of 100 pairs of input images randomly sampled from 10 ImageNet \cite{deng2009imagenet} classes. We train a separate Mood Space $M$ for each pair of input images, and interpolate between the two input images.
Additionally, we measure the perceptual similarity between pairs of consecutive images in the interpolation using LPIPS distance \cite{zhang2018unreasonable} and CLIP image similarity \cite{radford2021learning}.

In Table \ref{tab:eval_image_interpolation}, we report the maximum LPIPS between any consecutive pair of interpolated images (`Max LPIPS'),  the minimum CLIP similarity between any consecutive pair of images (`Min CLIP'), and the variance of LPIPS (`Var LPIPS') and the variance of CLIP similarity (`Var CLIP') across all consecutive pairs of images.

\begin{table}[h!]
\vspace{-5pt}
\centering
\resizebox{\columnwidth}{!}{%
\begin{tabular}{lccccc}
\toprule
    & \makecell{Max\\LPIPS $(\downarrow)$} & \makecell{Var\\LPIPS $(\downarrow)$} & \makecell{Min\\CLIP $(\uparrow)$} & \makecell{Var\\CLIP $(\downarrow)$} & \makecell{User\\Preference} \\
\midrule
Baseline  & 0.769  & 0.033  & 0.602  & 0.017  & 32\% \\
Ours      & 0.704  & 0.009  & 0.674  & 0.008  & 68\% \\
\bottomrule
\end{tabular}
}
\vspace{-5 pt}
\caption{Our method finds a smooth path connecting two visual concepts compared to the baseline linear interpolation. We measure the perceptual similarity and semantic similarity across pairs of consecutive images along the interpolation path.}
\label{tab:eval_image_interpolation}
\vspace{-5 pt}
\end{table}

Furthermore, we conduct a user preference study to assess how smoothly the generated images interpolate between the two inputs. We present the interpolation results from the baseline and our method, and we ask five people to rate which set of interpolated images is smoother. As shown in Table \ref{tab:eval_image_interpolation}, the user preferences indicate that our method provides a more gradual transition between images.

\subsection{Finding Paths for Consistent Composition}
\label{sec:eval_consistent_composition}

\begin{figure*}[!th]
    \centering
    \includegraphics[trim=0in 4.5in 3.8in 0in, clip,width=\textwidth]{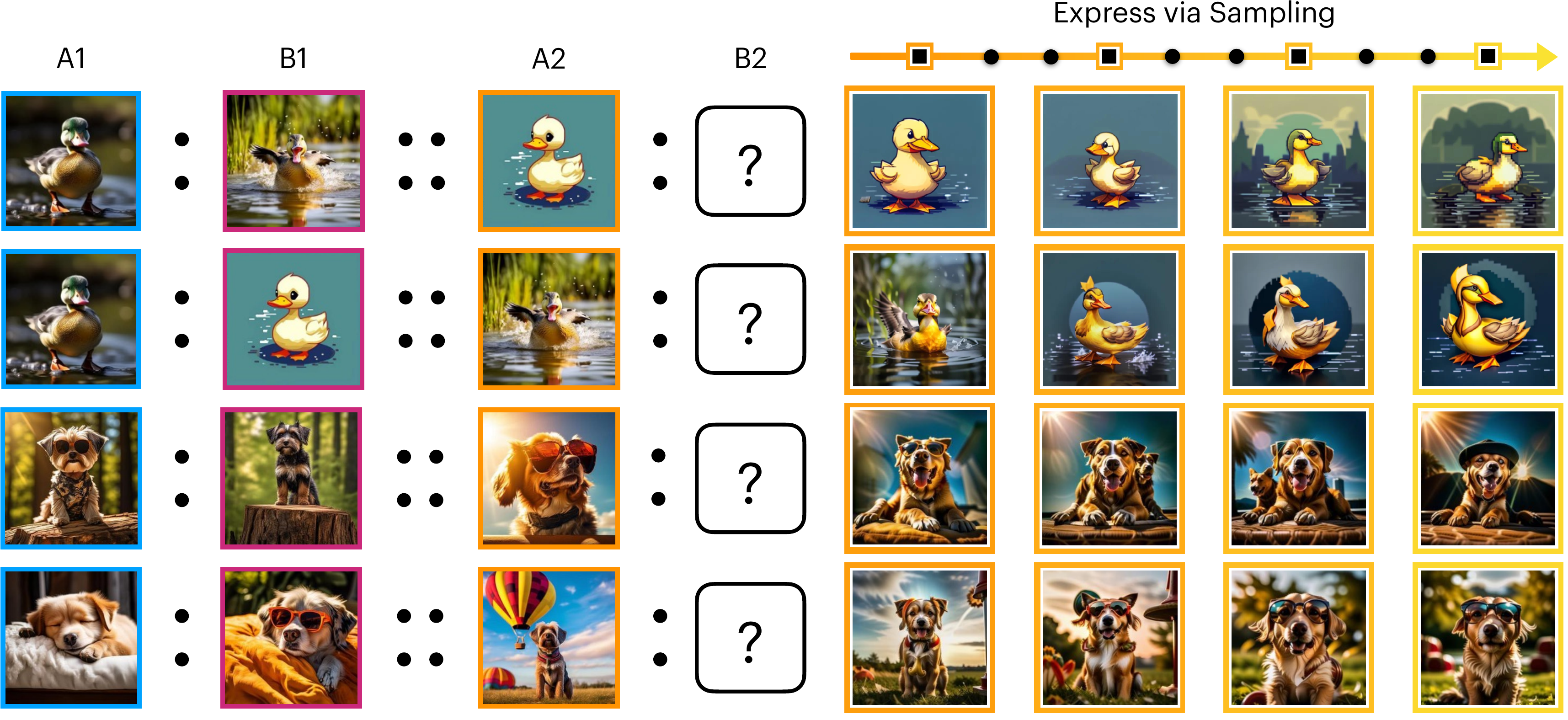}
    \caption{Consistency of visual analogy. Visual analogy can be achieved by lifting the path between A1 and B1 to A2, obtaining a sample of B2 that completes the analogy: A1 is to B1, as A2 is to B2. By switching the analogy pair from A1 : B1 to A1 : A2, we can create a different analogy: A1 is to A2, as B1 is to B2. \textbf{Top:} In the first case with the ducks, the connecting concept is the pose change. In the second case, the connecting concept is the style change. \textbf{Bottom:} In the first case with the dogs, the connecting concept is taking off glasses. In the second case, the connecting concept is wearing sunglasses.}
    \label{fig:visual_analogy}
    \vspace{-10 pt}
\end{figure*}

\textbf{Setup.} We evaluate whether different paths in our Mood Space lead to consistent outcomes through visual analogy tasks.
Given a connecting pair A1 : B1 lifted to another concept A2, we generate B2 to complete the analogy. Then, we test whether swapping A2 and B1, such that the new connecting pair is A1 : A2 is lifted to B1, produces consistent results, effectively measuring if different connecting and lifting paths yield similar outputs. If so, this indicates the integrability of paths in the Mood Space, which is a necessary condition for consistent composition.

\noindent \textbf{Quantitative Results.} We construct 50 analogy examples across five different types of visual transformations (e.g., photo-to-pixel-art, style transfer) using text-to-image generation \cite{chen2023pixart}. For each example, we:
\begin{itemize}
    \item Generate B2 using  original config (A1 : B1 :: A2 : B2)
    \item Generate B2' using  swapped config (A1 : A2 :: B1 : B2')
    \item Assess CLIP similarity and DreamSim between (B2, B2')
\end{itemize}

As shown in Figure \ref{fig:visual_analogy}, our method successfully maintains consistency across different paths. Table \ref{tab:eval_visual_analogy} demonstrates high similarity between outputs generated via different paths, confirming that our Mood Space enables consistent visual analogies regardless of the chosen path.

\begin{table}[!ht]
\centering
\resizebox{0.8\columnwidth}{!}{%
\begin{tabular}{lccc}
\toprule
    & Mean CLIP $(\uparrow)$ & Mean DreamSim $(\downarrow)$ \\ \midrule
Baseline   &  0.713  & 0.559     \\ 
Ours       &  0.756  & 0.454     \\
\bottomrule
\end{tabular}
}%
\caption{Evaluating the composition consistency of visual analogies. Higher consistency indicates a closer approximation to path integrability in the Mood Space.}
\label{tab:eval_visual_analogy}
\vspace{-10 pt}
\end{table}

\subsection{Ablation Study}
\label{sec:ablation_study}

\begin{figure}
    \centering
    \includegraphics[width=1\linewidth]{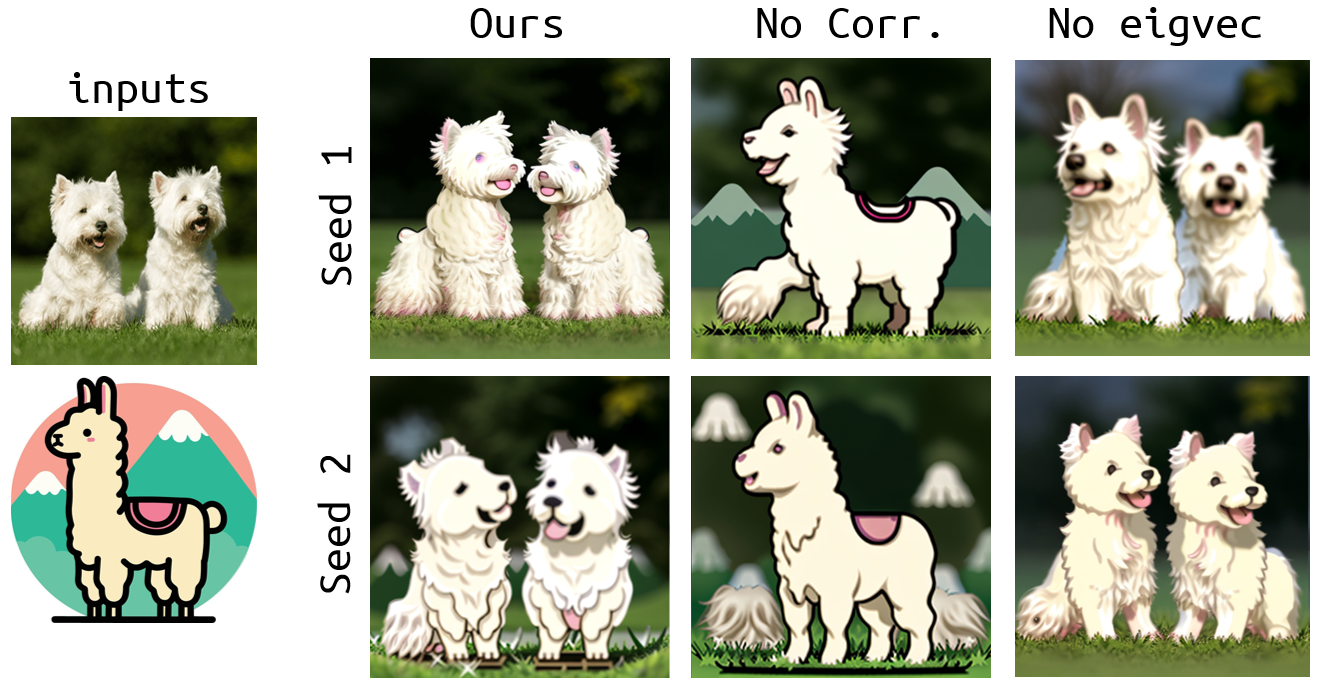}
    \caption{Ablation study on DINO correspondence and spectral loss. Without DINO correspondence, the object count is not preserved. Without spectral loss, the concepts are not connected in the Mood Space, indicated by the head having a dog style while the body has a llama style.}
    \label{fig:ablation}
\end{figure}

First, we want to study the effect of DINO correspondence on the generated image. As shown in Figure \ref{fig:ablation}, we chose a case with two dogs in the source image and one llama in the target image. Without DINO correspondence, only one dog or llama is generated. Using our approach, we maintain the count of two animals that are a hybrid dog-llama.

Second, we analyze the effect of the spectral loss, which aims to bring out and connect the relevant features across the two images. Without spectral loss, the head of the generated animal is a dog while the body is a llama, which means that the dog and llama concepts are disconnected.
in the Mood Space.

\begin{figure*}[]
    \centering
    \includegraphics[trim=0in 4.9in 0.5in 0in, clip,width=\textwidth]{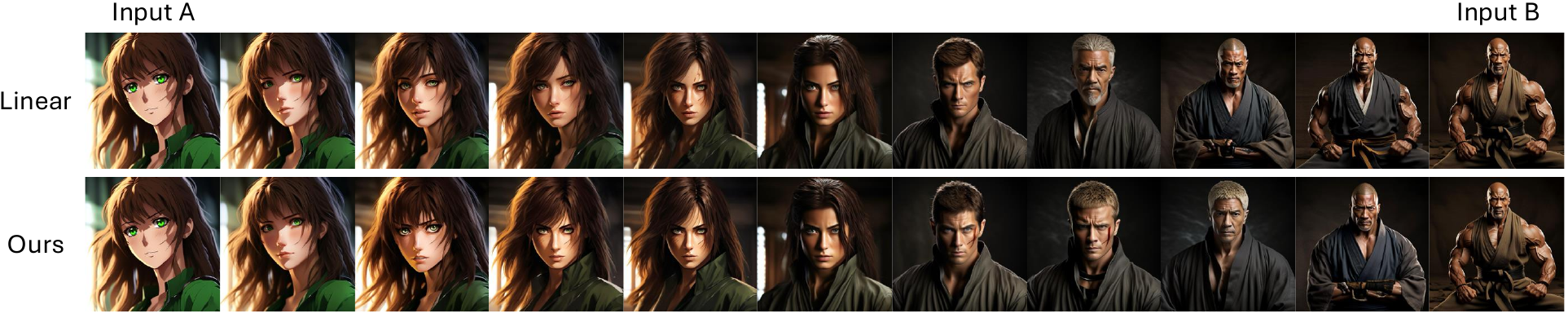}
    \vspace{-20 pt}
    \caption{We extend our Mood Space to text encoding. Using 700 prompts from the DiffusionDB dataset as the context set, our interpolation between two prompts leads to more gradual change in head motion, hair style, and appearance compared to the baseline linear interpolation.}
    \label{fig:text_interpolation}
    \vspace{-10pt}
\end{figure*}

\begin{figure}
    \centering
    \vspace{-8pt}
    \includegraphics[width=1\linewidth]{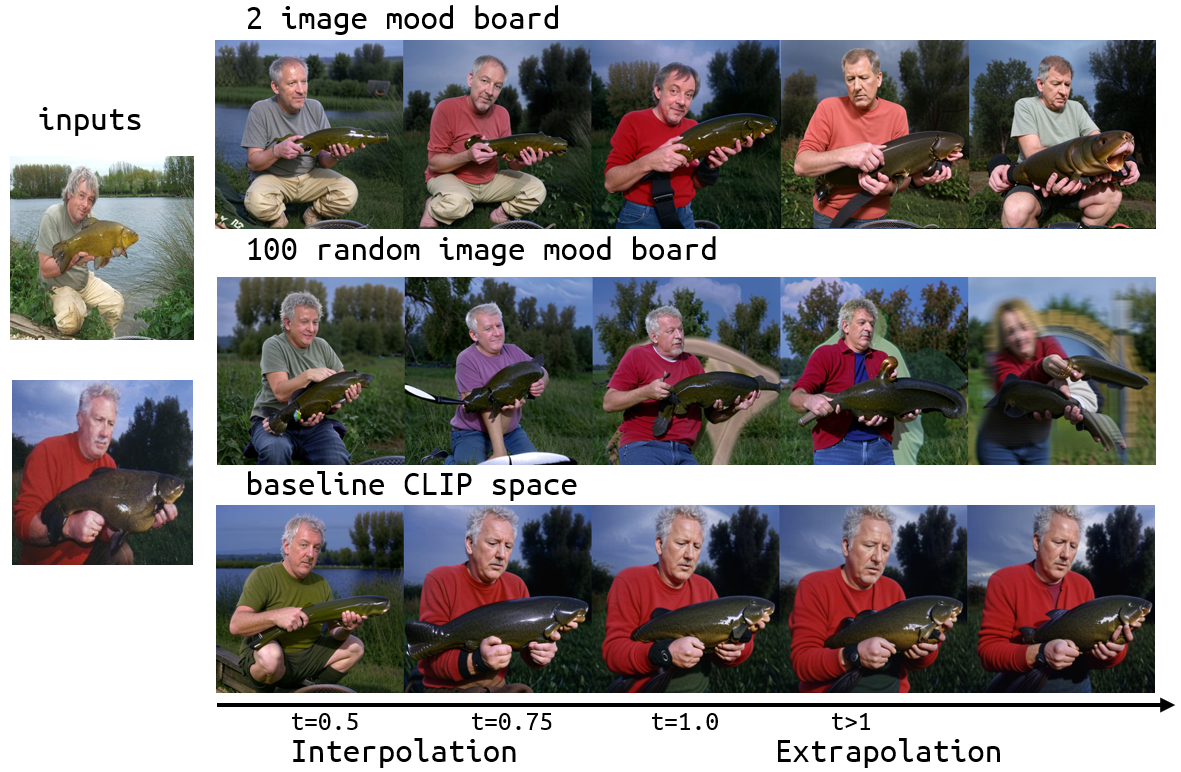}
    \caption{We analyze the effect of interpolation and extrapolation in the Mood Space, under the condition with and without distracting objects in the Mood Board. \textbf{Top Row:} Using two context images produces excellent extrapolation results. \textbf{Middle Row:} Using 100 random context images produces decent interpolation but more random extrapolation results. \textbf{Bottom Row:} Extrapolating in the original CLIP space fails.}
    \label{fig:analysis_context}
    \vspace{-12pt}
\end{figure}

\begin{table}[h!]
\vspace{-5pt}
\centering
\resizebox{\columnwidth}{!}{%
\begin{tabular}{lccccc}
\toprule
    & Max LPIPS $(\downarrow)$ & Max DreamSim $(\downarrow)$ \\
\midrule
Ours w/o DINO matching &  0.633  &  0.434  \\
Ours                   &  0.627  &  0.395  \\
\bottomrule
\end{tabular}
}
\caption{Effect of DINO correspondence on the structural similarity between the first input image and each interpolated image, such as object count, shape, and position.}
\label{tab:ablation_dino_correspondence}
\end{table}

\subsection{Analysis}
\label{sec:analysis}

\textbf{Uniformity of Embedding Space.}
The embedding space should be evenly distributed to ensure that pairwise interpolation between any pair of points has nearby samples.  This amounts to the entropy of the embedding space being high. We randomly sample 1000 images from ImageNet and train a Mood Space compression model using an estimated intrinsic dimension of $G=22$. Then, we use Shannon Entropy (normalized to 0 to 1) on the image embeddings.

To be more fair, we use PCA to project the CLIP and DINO features to a lower dimension (250), low enough to preserve the information. We then measure entropy in this compressed PCA space. We look at the distribution of eigenvalues in this PCA compression. Ideally, we want similar eigenvalues, indicating a compact space. We measure the entropy of the eigenvalues.

\begin{table}[h]
    \centering
    \resizebox{1.0\columnwidth}{!}{%
        \begin{tabular}{lccc}
            \toprule
            & \makecell{DINO} & \makecell{CLIP} & \makecell{Mood Space \qquad}  \\
            \midrule
            Entropy            & 0.717 \quad & 0.465 \quad & 0.755  \qquad \\
            Entropy (PCA-ed)   & 0.648 \quad & 0.519 \quad & 0.711 \quad \\
            Entropy (PCA eigvals) \qquad   & 0.506 \quad & 0.432 \quad & 0.957 \quad \\
            \bottomrule
        \end{tabular}%
    }
    \vspace{-6pt}
    \caption{Measuring the uniformity of the embedding space using entropy values for DINO, CLIP, and Mood Space. We want a higher entropy value.}
    \label{tab:entropy}
\end{table}

\noindent \textbf{Effect of Number of Concepts on Mood Space.}
We examine the interpolation and extrapolation results to probe the quality of the Mood Space beyond the two reference objects. We observe that distracting objects in the context set does not affect the interpolation quality significantly but leads to poor quality extrapolation to one of the objects in the distracting set.
As shown in Figure \ref{fig:analysis_context}, using two well-curated context images surprisingly achieves excellent extrapolation. As a sanity check, we verified that extrapolation in the original CLIP embedding space leads to similar results as the interpolation.

\noindent \textbf{Connecting Concepts in Natural Language.}
We adapt our Mood Space to find connections between distinct concepts in the text embedding space. We extract 20 out of 120 text tokens using 700 context prompts from the DiffusionDB dataset, and generate the final image using PixArt \cite{chen2023pixart}. In Figure \ref{fig:text_interpolation}, we use our Mood Space to interpolate between two sets of text tokens, and compare with a baseline linear interpolation. We observe that traversing between the text embeddings in Mood Space results in a more smooth transition in head motion, hair style, and appearance.

\section{Conclusion}
\label{sec:conclusion}

We present Mood Space, a novel approach for learning a semantically meaningful and controllable latent space for image manipulation. Our method learns a compressed representation that brings the relevant features between context images closer, thus enabling smooth interpolation between visual concepts.  The Mood Space also provides consistent results when traversing different paths, and maintains local image structure.   Learning is efficient, using two simple 4-layer token-wise MLPs, with few context images.  Through experiments, we demonstrated that our method generates higher quality results than direct interpolation in existing embedding spaces, achieving better smoothness metrics and semantic preservation while maintaining path consistency in visual analogy tasks. 
While the method works best with a focused Mood Board containing a limited number of related concepts and excels primarily at interpolation rather than extrapolation, we believe Mood Space opens up new possibilities for intuitive and controllable image manipulation, providing a foundation for future research in semantic image editing and creative visual tools.

\textbf{Acknowledgements.}
This work is supported by funds provided by the National Science Foundation and by DoD OUSD (R\&E) under Cooperative Agreement PHY-2229929 (The NSF AI Institute for Artificial and Natural Intelligence).

{
    \small
    \bibliographystyle{ieeenat_fullname}
    \bibliography{main}
}

\clearpage
\setcounter{page}{1}
\maketitlesupplementary

\begin{appendix}
\section{Additional Method Details}

\subsection{Token Path Lifting}

We take inspiration from fiber bundles. If we have a fiber bundle $\pi_W: W\to M$, we can take a path $\gamma(t): [0,1]\to M$ from $m_{A_1}=\gamma(0)$ to $m_{A_2}=\gamma(1)$. Furthermore, we assume we have some point $w_{B_1}$ in the fiber $\pi_W^{-1}(m_{A_1})$, in other words $\pi_W(w_{B_1})=m_{A_1}$. Then we say $\hat{\gamma}:[0,1]\to W$ is a lifting of $\gamma$ starting at $w_{B_1}$ if $\hat{\gamma}(0)=w_{B_1}$ and $\pi_W(\hat{\gamma}(t))=\gamma(t)$. One of the key properties of fiber bundles, and fibrations more generally is that paths in the base space can have liftings starting at each point in the fiber above $\gamma(0)$. 

Our idea is that when we have image tokens $u_{A_1}$ and $u_{A_2}$, we can project these points into the Mood Space $M$ via the projection map $\pi_V$, connect them in Mood Space with a path $\gamma$, then lift this path starting at the embedding of another token $w_{B_1}$ along $\hat{\gamma}$. This allows us to transport the token $w_{B_1}$ into a novel token $w_{B_3}:= \hat{\gamma}(1)$. 

In this work, we have constructed the section $\sigma_W$ precisely for the purposes of lifting paths. Here we take a naive path in $M$, $\gamma(t) = m_{A_1} + t*(m_{A_2}-m_{A_1})$ exploring the fact that we have constructed our Mood Space $M$ as a vector space. Since $V$ is also a vector space we could construct $\hat{\gamma}(t) = w_{B_1} + t*\sigma_v( m_{A_2}-m_{A_1})$. Since we don't actually have a bundle map $\pi_W$, we are simply using the notion of lifting the path, as guidance. We note that we could have taken a sequence of points $t_1=0\leq t_2\leq \ldots \leq t_Q=1$, and lift each segment of $\gamma$ between $t_k$ and $t_{k+1}$ via $\sigma_W$ to achieve path in W that more closely follows $\gamma([0,1])\subset M$.

\subsection{From Tokens to Images}

We want to promote the token path lifting to more of an image path lifting. Given images $A_1$, $A_2$, $B_1$, we construct a path from $\pi_V(A_1)$ to $\pi_V(A_2$) based on our path in Mood Space, and essentially starting at $B_1$ to determine $B_2$ in $W$. To construct this path we use the Ncut framework to cluster all tokens. This allows us to cluster the tokens into $H$ token-groups. When considering the $256 = 16\times 16$ tokens that make up image $A_1$, we group them into the $H$ token groups. We do the same thing for $A_2$ and $B_1$. For each of the $H$ groups of $A_1$, and separately $A_2$ we compute the centroid of the tokens in that group. So for $A_1$ we have the $C_{A_1,i}$ for $1\leq i \leq H$ and similarly for $A_2$. 

Next, we compute a correspondence between the token group centroids to align $C_{A_1,i}$ with corresponding token group centroids $C_{A_2,j}$ where $j = P(i)$ and $P:\{1,\ldots, H\} \to \{1,\ldots, H\}$ determined by maximizing over $j$, the token affinity $F$ between $C_{A_1,i}$ and $C_{A_1,j}$.

Next we create a path $\gamma_i(t)$ in Mood Space from $\pi_V(C_{A_1,i})$ to $\pi_V(C_{A_2,P(i)})$ for each $1\leq i \leq H$. Then for each embedded token in $v_{B_1}=E_V(B_1)$ we use the token clustering to identify the token group it belongs, lets say group $i$ and we lift $\gamma_i$ starting at $w_{B_1}=\sigma_W(\pi_V(v_{B_1})=\hat{\gamma}(0)$ and define $w_{B_1}:= \hat{\gamma}(1)$.

\section{DINO Correspondence}

\begin{figure}[h!]
    \centering
    \includegraphics[width=1\linewidth]{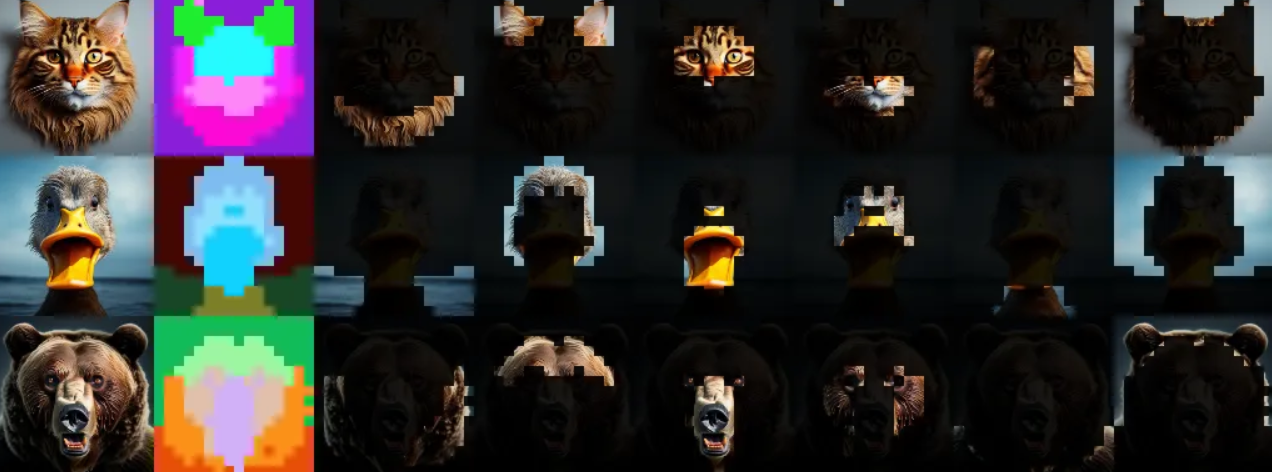}
    \caption{In Token Path Lifting, we use DINO feature to do token clustering and correspondence matching. Token clusters from the 3 images are matched during token path lifting. Each column shows cluster in correspondence.}
    \label{fig:dino_matching}
\end{figure}

To construct the token clusters, we use the discrete clustering \cite{1238361} with DINO feature input. We cluster all tokens in each image separately. When considering the spatial tokens $256 = (16\times 16)$ tokens that make up image $A_1$, we group them into the $H=10$ token groups. We do the same thing for $A_2$. For each of the $H$ groups of $A_1$, and separately $A_2$ we compute the feature centroid (in DINO space) of the tokens in that group. So for $A_1$, we have the $C_{A_1,i}$ for $1\leq i \leq H$ and similarly for $A_2$. 

Next, we compute a correspondence between the token group centroids to align $C_{A_1,i}$ with corresponding token group centroids $C_{A_2,j}$ where $j = P(i)$ and $P:\{1,\ldots, H\} \to \{1,\ldots, H\}$ determined by maximizing over $j$ the token affinity $F$ between $C_{A_1,i}$ and $C_{A_1,j}$.

Next we create a path $\gamma_i(t)$ in Mood Space from $\pi_V(C_{A_1,i})$ to $\pi_V(C_{A_2,P(i)})$ for each $1\leq i \leq H$. 

\section{Implementation Details}

\subsection{Hyperparameters}

\paragraph{Mood Space MLP Model.}

In Section 3 of the main text, we train an encoder MLP, denoted as $\pi_{V,\Theta}$, and a decoder MLP, denoted as $\sigma_{W,\Theta'}$. Both networks consist of four layers with 512 hidden units, resulting in approximately one million parameters.

\paragraph{Regularization Loss Balancing.}

The training loss function is defined as a weighted combination of multiple loss terms, including spectral graph embedding loss, Riemannian curvature loss, repulsive force loss, reconstruction loss, and covariance loss:

\[
\begin{aligned}
L(\Theta, \Theta') &= L_{\text{spec}}(\Theta) + \lambda_1 L_{\text{curv}}(\Theta) + \lambda_2 L_{\text{rep}}(\Theta) \\
&\quad + \lambda_3 L_{\text{recon}}(\Theta, \Theta') + \lambda_4 L_{\text{var}}(\Theta).
\end{aligned}
\]

The specific values of the regularization coefficients are provided in Table~\ref{tab:regularization}. Empirical results indicate that setting $\lambda_1, \lambda_2, \lambda_4$ to $1 \times 10^{-5}$ ensures that the corresponding regularization terms remain significant in the optimization process.

\begin{table}[h]
\centering
\caption{Regularization Loss Coefficients}
\label{tab:regularization}
\begin{tabular}{c|c}
\hline
\textbf{Coefficient} & \textbf{Value} \\ 
\hline
$\lambda_1$ (Curvature) & $1 \times 10^{-5}$ \\  
$\lambda_2$ (Repulsion) & $1 \times 10^{-5}$ \\  
$\lambda_3$ (Reconstruction) & $1$ \\  
$\lambda_4$ (Covariance) & $1 \times 10^{-5}$ \\  
\hline
\end{tabular}
\end{table}

\paragraph{Optimizer and Training.}

The model is trained using the Adam optimizer with a learning rate of $1 \times 10^{-3}$. The number of training steps required for convergence varies depending on the dataset size:

\begin{table}[h]
\centering
\caption{Training Convergence, two-image training converges in within 1 minute on a RTX4090 GPU.}
\label{tab:training}
\begin{tabular}{c|c}
\hline
\textbf{Experiment} & \textbf{Convergence Steps} \\ 
\hline
Two-image Mood Space & 1,000 steps \\  
One hundred-image Mood Space & 10,000 steps \\  
\hline
\end{tabular}
\end{table}

\subsection{Affinity and Spectral Graph Embedding}

It is important to note from a computational point of view, that explicitly computing the eigenvalues and the affinity maps quickly becomes prohibitively expensive. However, when the final measures we need to compute are obtained, they can be well estimated by sampling. We use farthest point sampling (FPS) to select a fixed-size subset of 512 nodes.

For the Spectral Graph Embedding loss, as described in Algorithm 1, we use a maximum of $k = 32$ eigenvectors. The sliding vector similarity mechanism ensures that the choice of $k$ remains robust. Since eigenvectors exhibit a hierarchical structure, selecting a smaller $k$ allows the loss function to focus on the most relevant features.

\begin{figure*}
    \centering
    \includegraphics[width=0.75\linewidth]{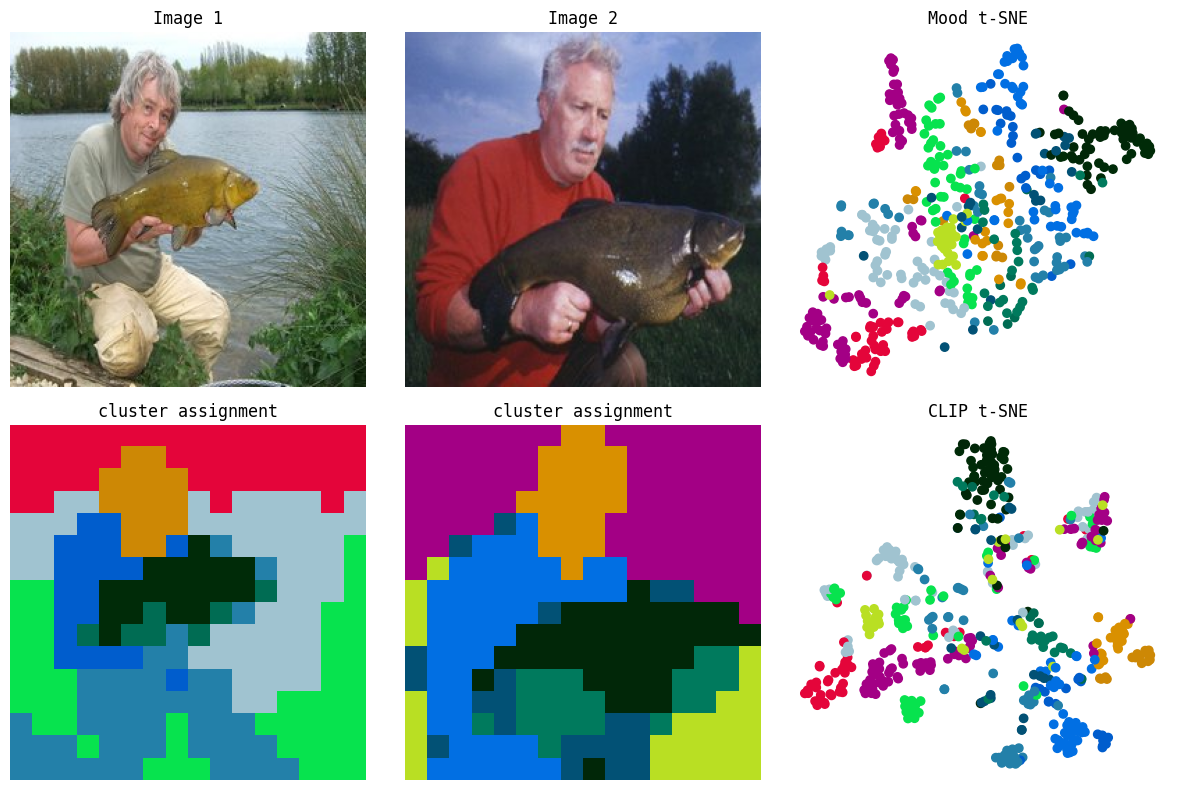}
    \caption{Image tokens in the Mood Space is more evenly distributed and connected than the original CLIP embedding space. This plot is t-SNE embedding of $2\times 256$ tokens from two images in Mood Space and CLIP space, in the Mood Space, tokens are more evenly distributed thus sampling on Mood Space is better than CLIP space.}
    \label{fig:enter-label}
\end{figure*}

\begin{figure*}
    \centering
    \includegraphics[width=0.5\linewidth]{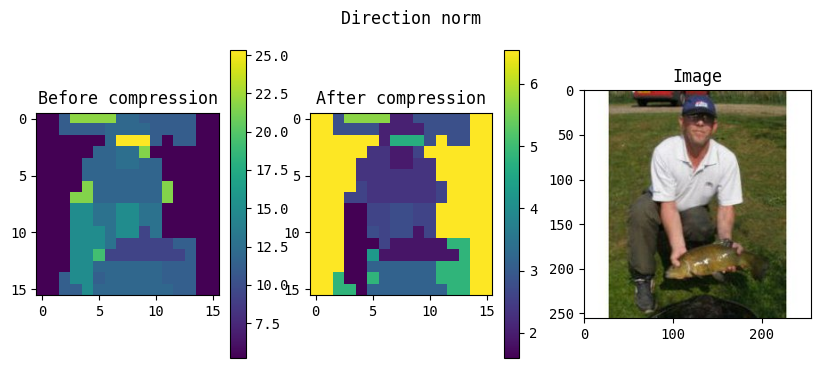}
    \caption{Mood Space compression bring in closer common relevant foreground object across two context images. Left: token similarity using CLIP feature, note background has smaller distance while foreground has larger distance. Middle: Mood Space bring in closer the foreground object while the background difference is amplified. This demonstrates the compression in Mood Space is able to remove the irrelevant feature while bringing in closer the relevant ones.}
    \label{fig:enter-label}
\end{figure*}

\begin{figure*}
    \centering
    \includegraphics[width=0.5\linewidth]{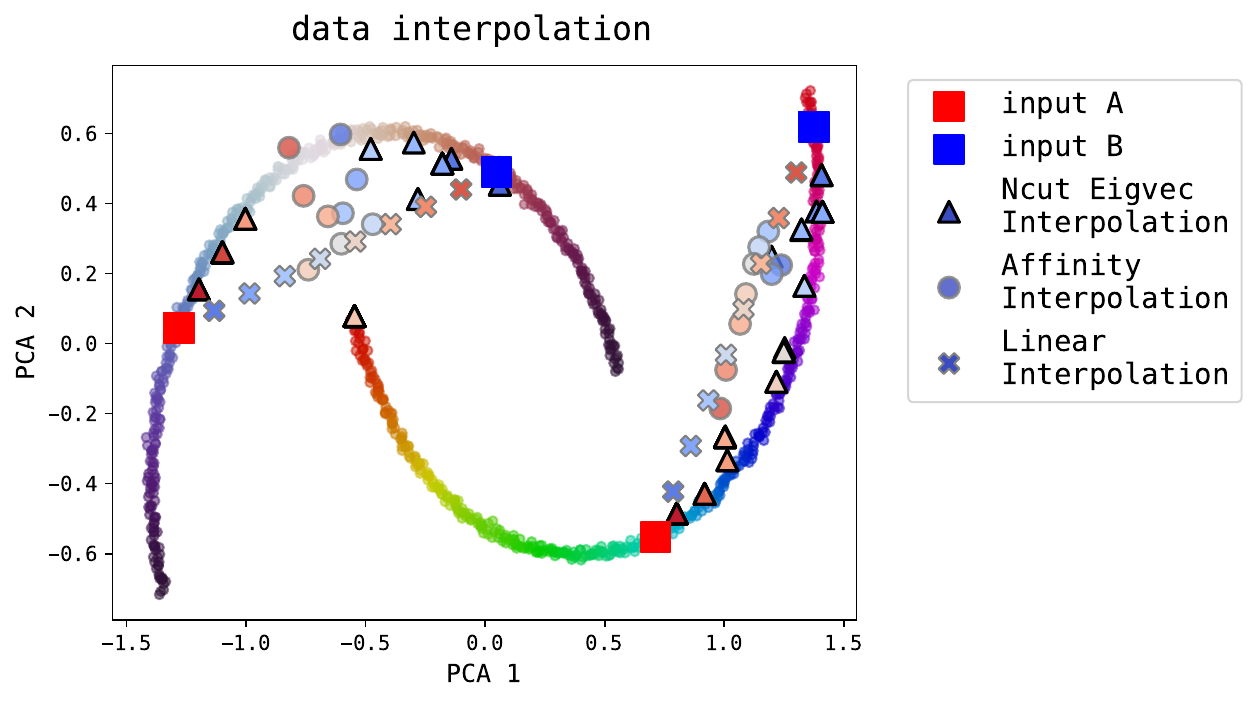}
    \caption{We use our Mood Space to interpolate a 2D toy example, and compare with a baseline 1) linear interpolation 2) use affinity loss instead of spectral graph embedding loss. We observe that interpolation in Mood Space (triangle, Ncut Eigvec) follow the manifold of data.}
    \label{fig:enter-label}
\end{figure*}

\clearpage
\begin{figure*}
    \centering
    \includegraphics[width=\linewidth]{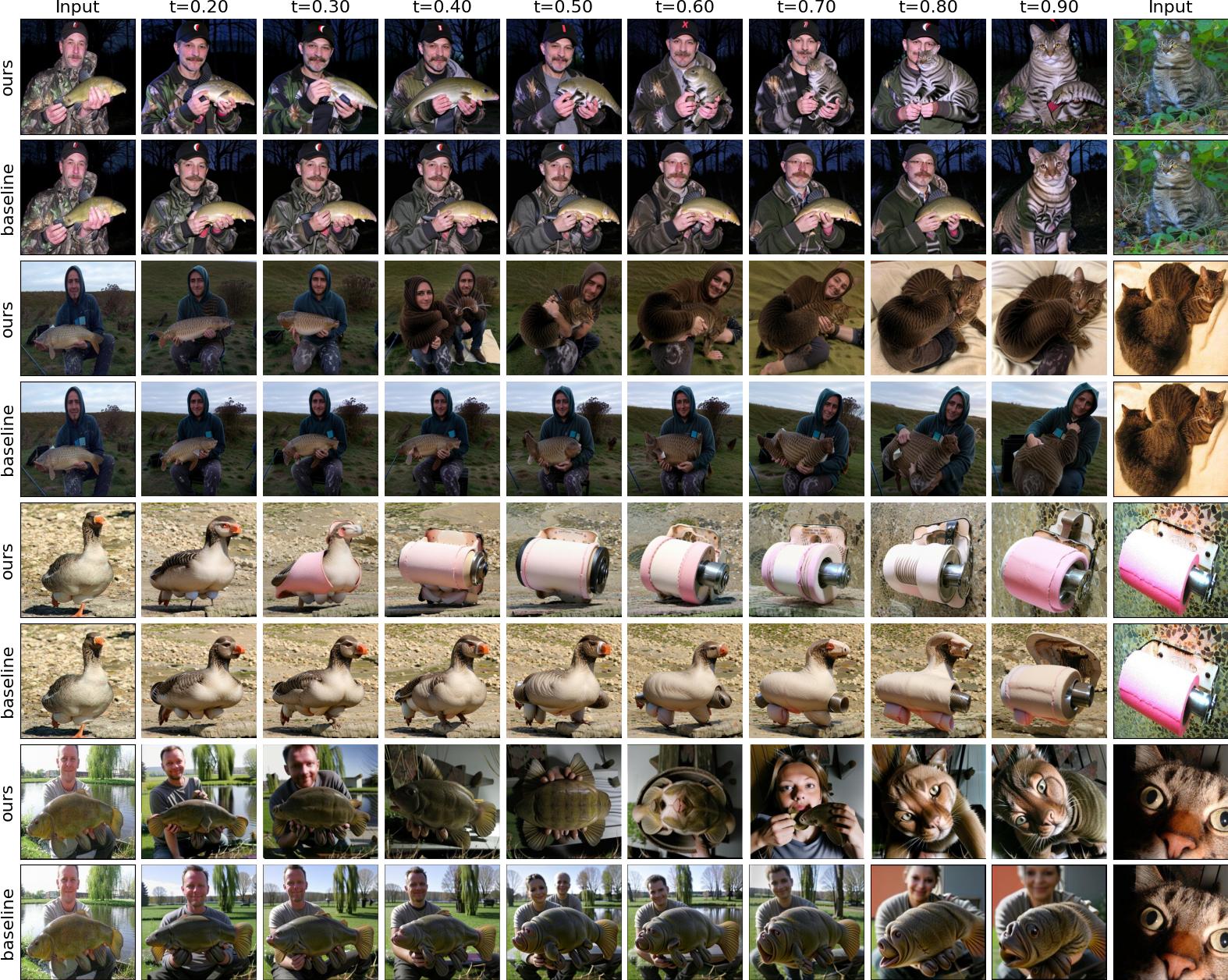}
    \caption{Interpolation results on Mood Space vs Original CLIP embedding space (baseline).}
    \label{fig:appendix_interpolation}
\end{figure*}

\clearpage
\begin{figure*}
    \centering
    \includegraphics[width=\linewidth]{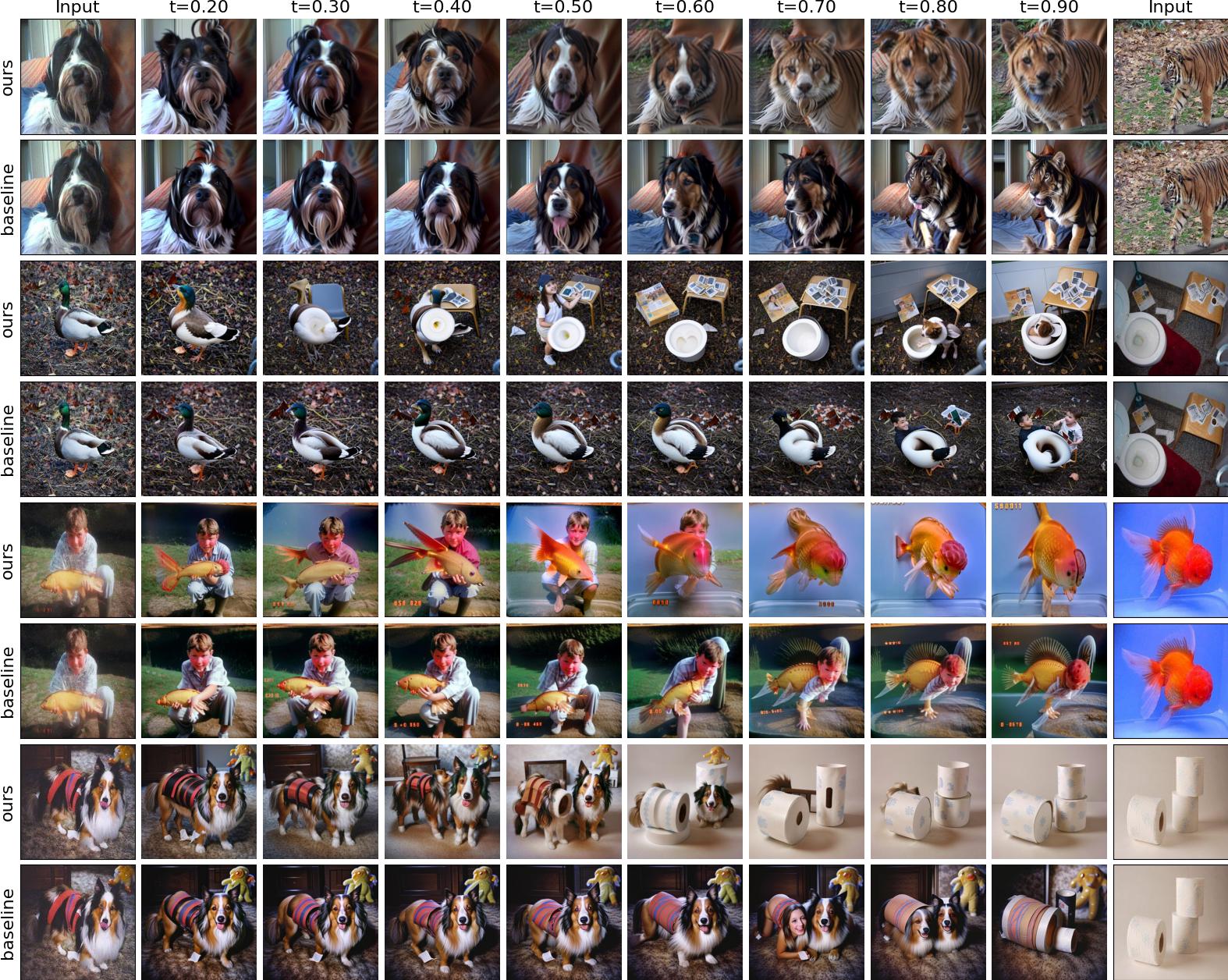}
    \caption{Interpolation results on Mood Space vs Original CLIP embedding space (baseline).}
    \label{fig:appendix_interpolation}
\end{figure*}

\clearpage
\begin{figure*}
    \centering
    \includegraphics[width=\linewidth]{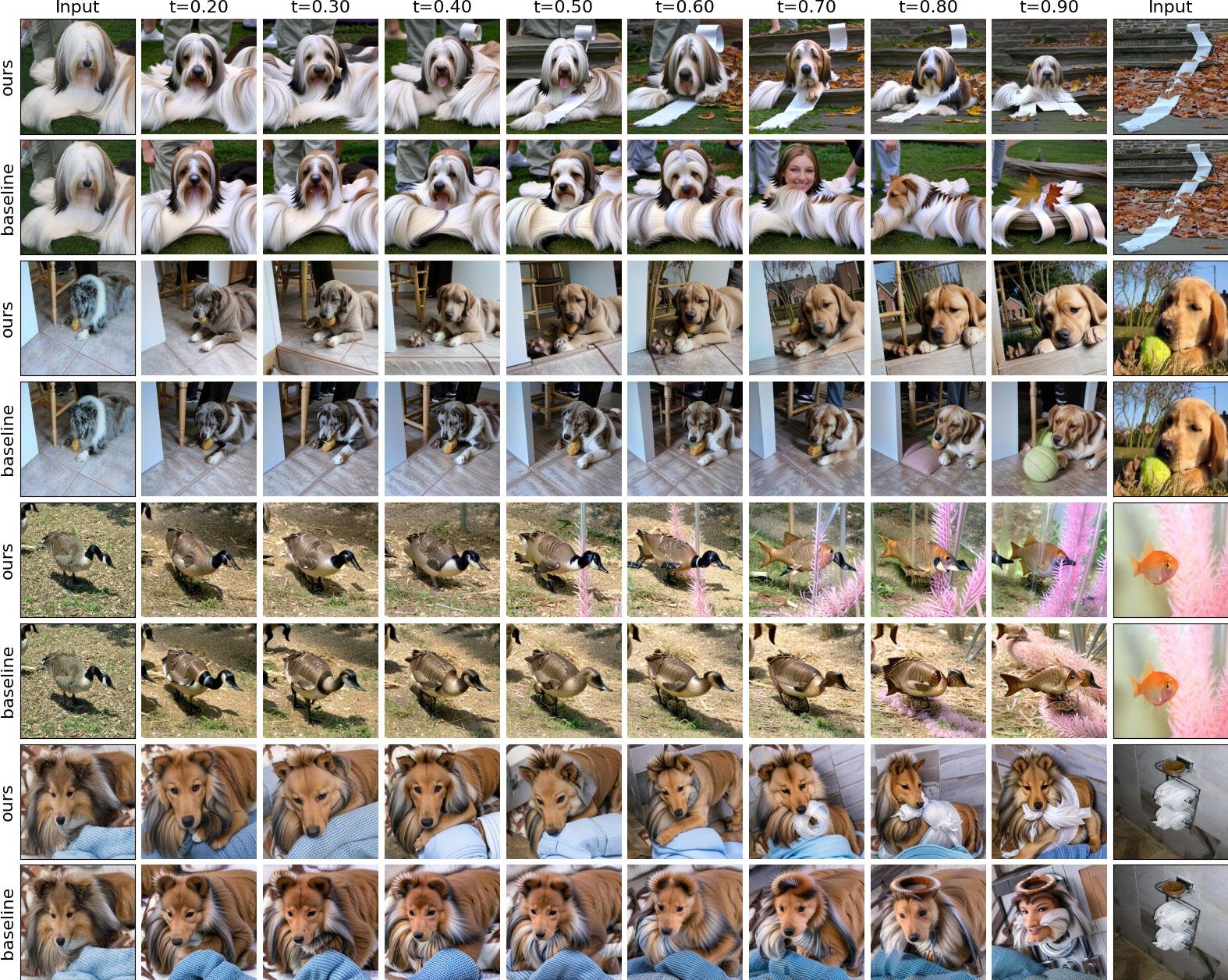}
    \caption{Interpolation results on Mood Space vs Original CLIP embedding space (baseline).}
    \label{fig:appendix_interpolation}
\end{figure*}

\clearpage
\begin{figure*}
    \centering
    \includegraphics[width=\linewidth]{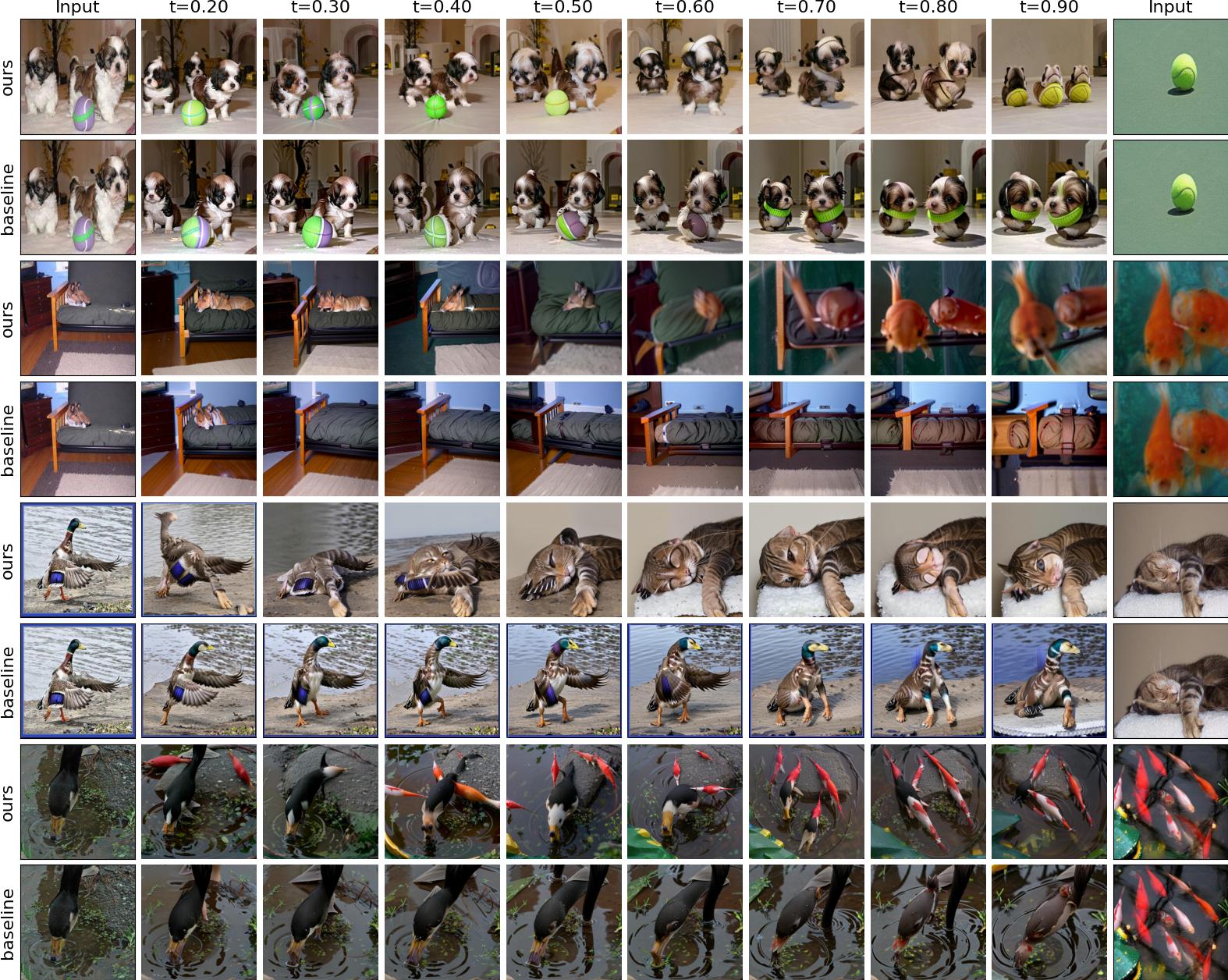}
    \caption{Interpolation results on Mood Space vs Original CLIP embedding space (baseline).}
    \label{fig:appendix_interpolation}
\end{figure*}

\clearpage
\begin{figure*}
    \centering
    \includegraphics[width=\linewidth]{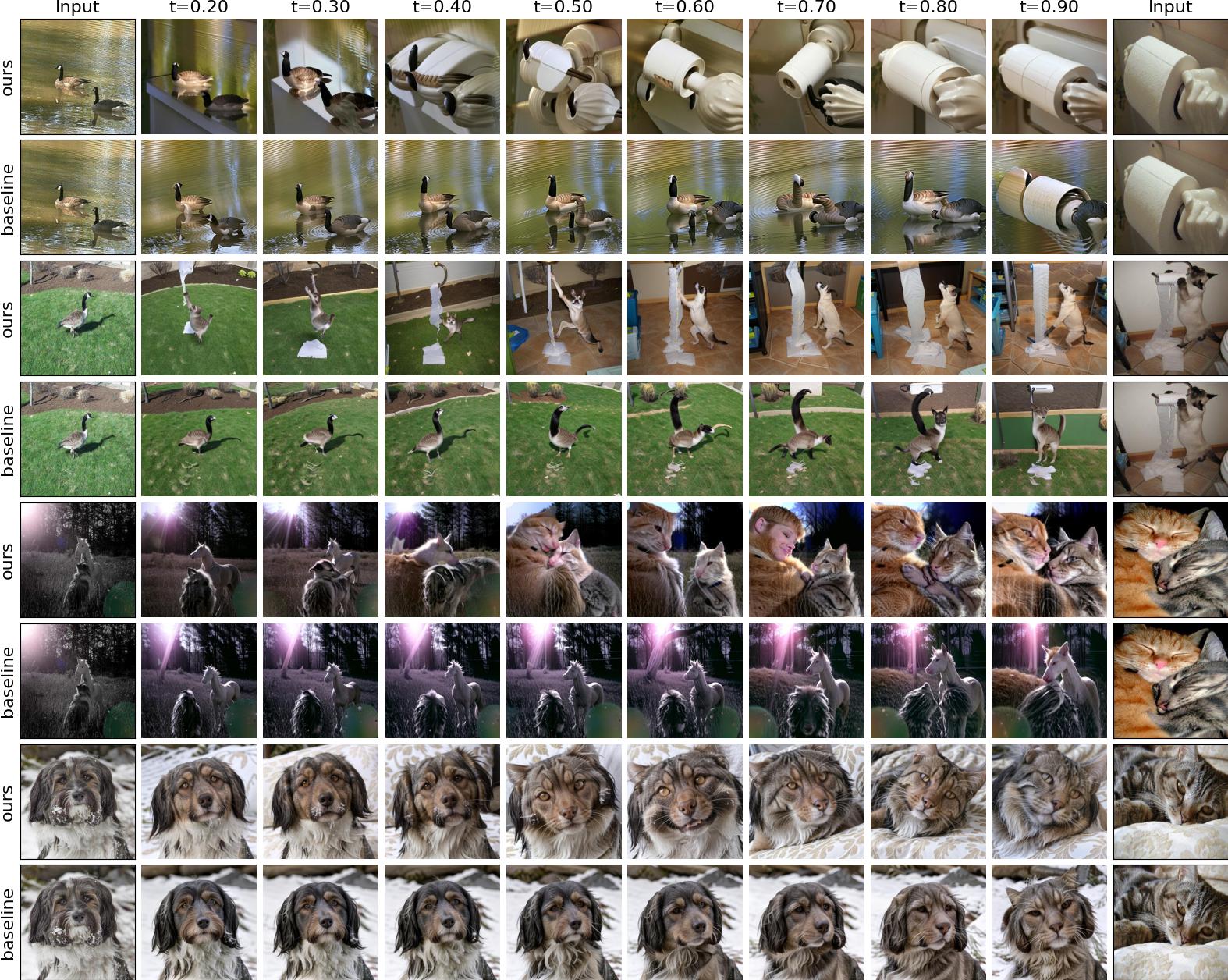}
    \caption{Interpolation results on Mood Space vs Original CLIP embedding space (baseline).}
    \label{fig:appendix_interpolation}
\end{figure*}

\clearpage
\begin{figure*}
    \centering
    \includegraphics[width=\linewidth]{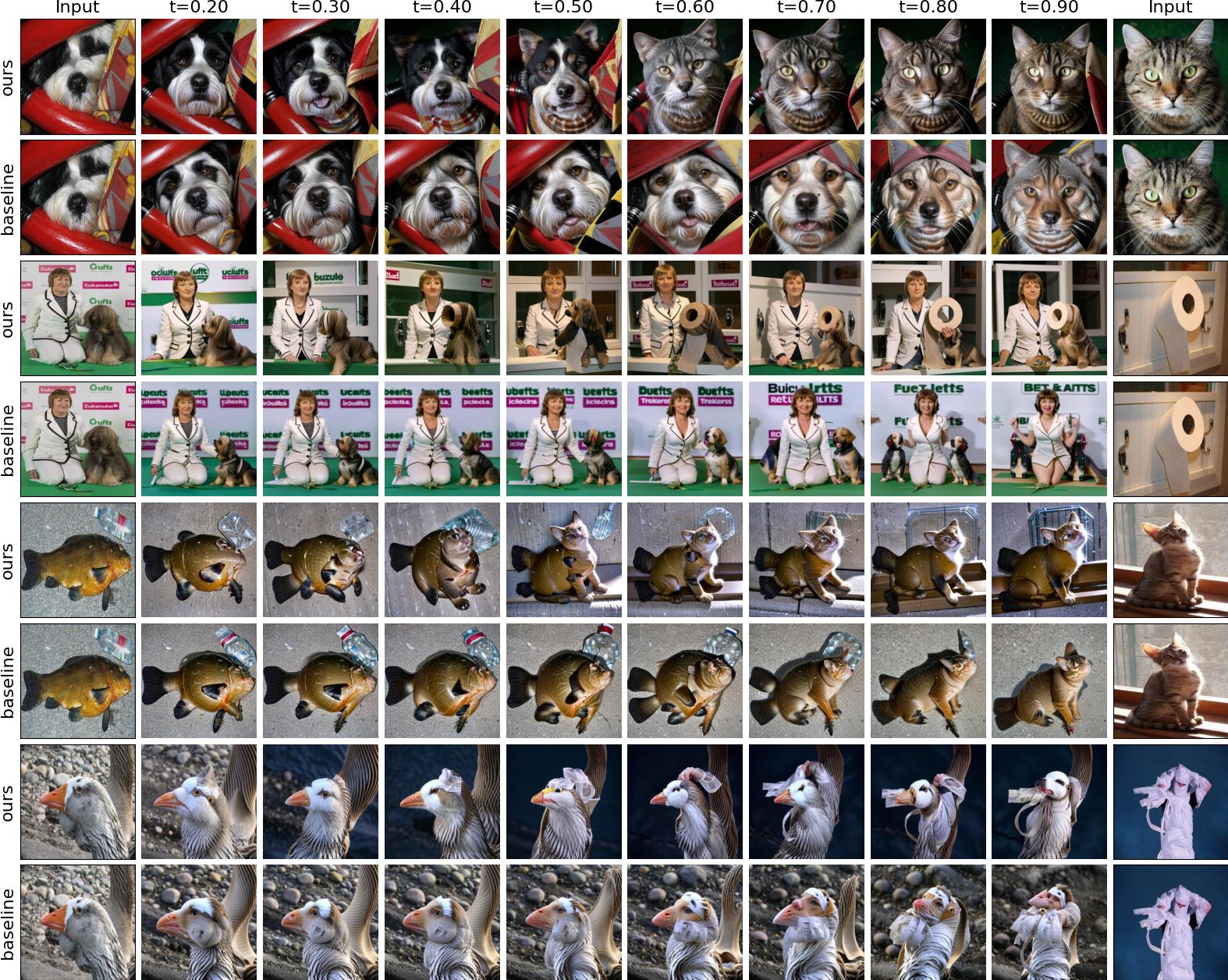}
    \caption{Interpolation results on Mood Space vs Original CLIP embedding space (baseline).}
    \label{fig:appendix_interpolation}
\end{figure*}

\clearpage
\begin{figure*}
    \centering
    \includegraphics[width=\linewidth]{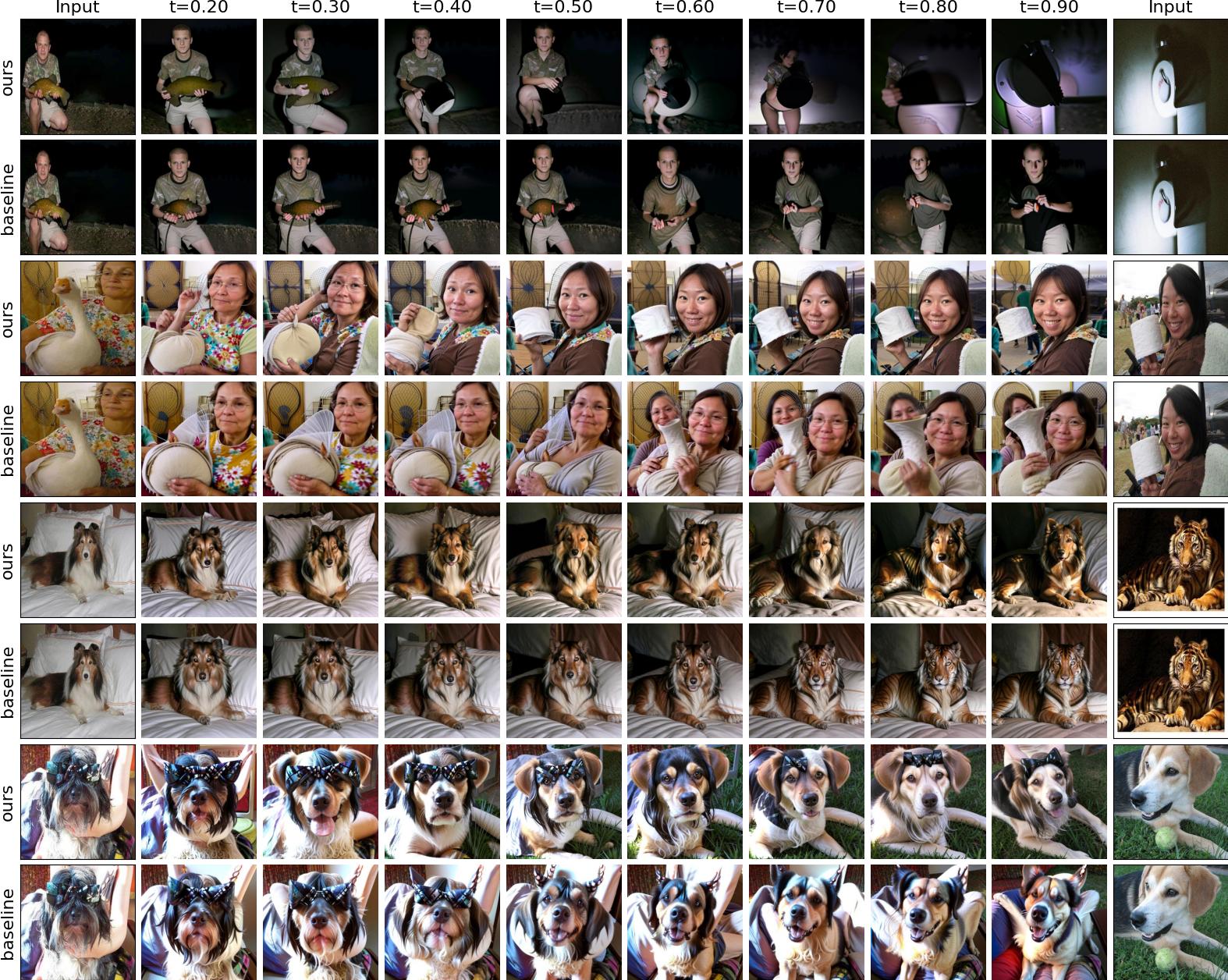}
    \caption{Interpolation results on Mood Space vs Original CLIP embedding space (baseline).}
    \label{fig:appendix_interpolation}
\end{figure*}

\clearpage
\begin{figure*}
    \centering
    \includegraphics[width=\linewidth]{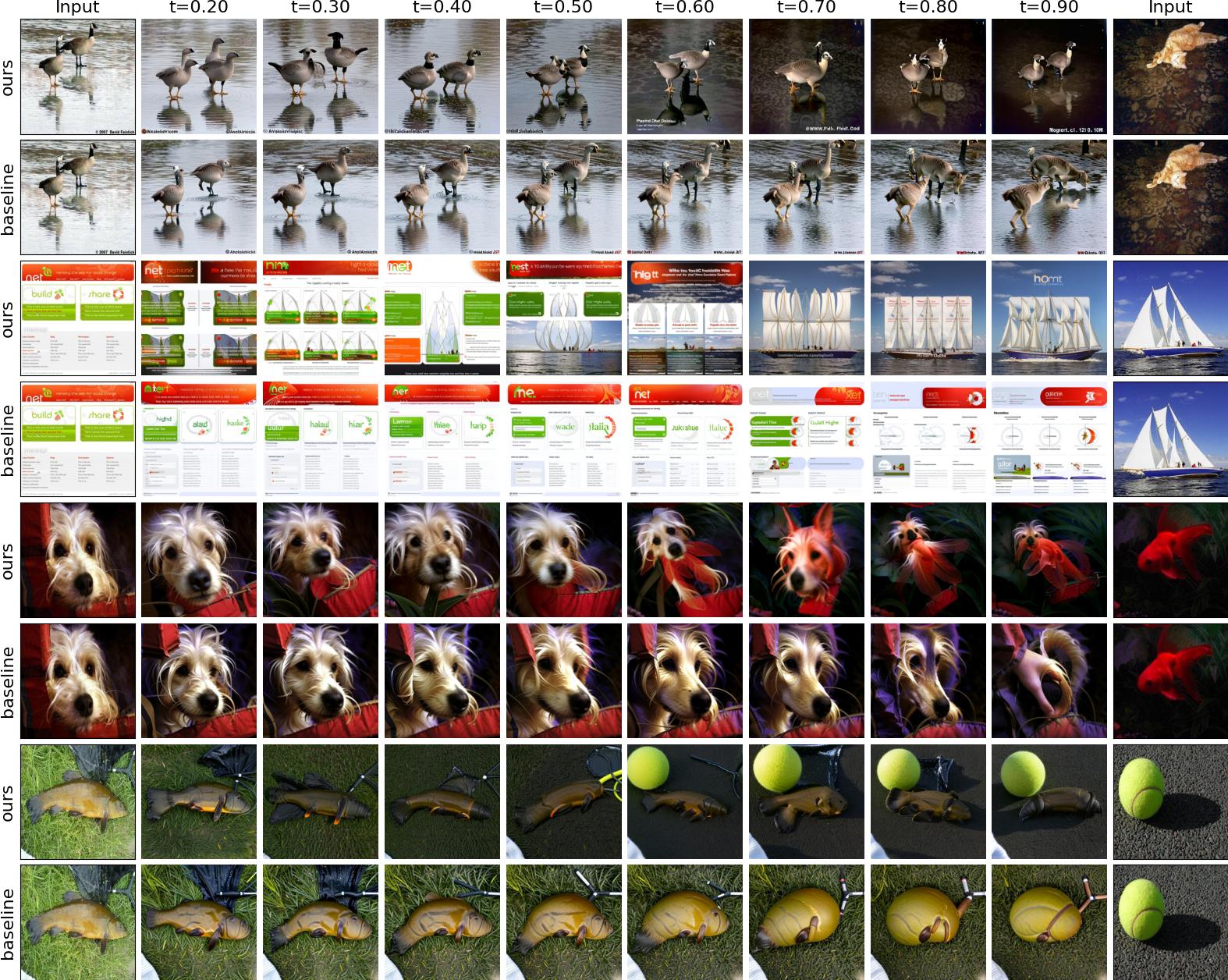}
    \caption{Interpolation results on Mood Space vs Original CLIP embedding space (baseline).}
    \label{fig:appendix_interpolation}
\end{figure*}

\clearpage
\begin{figure*}
    \centering
    \includegraphics[width=\linewidth]{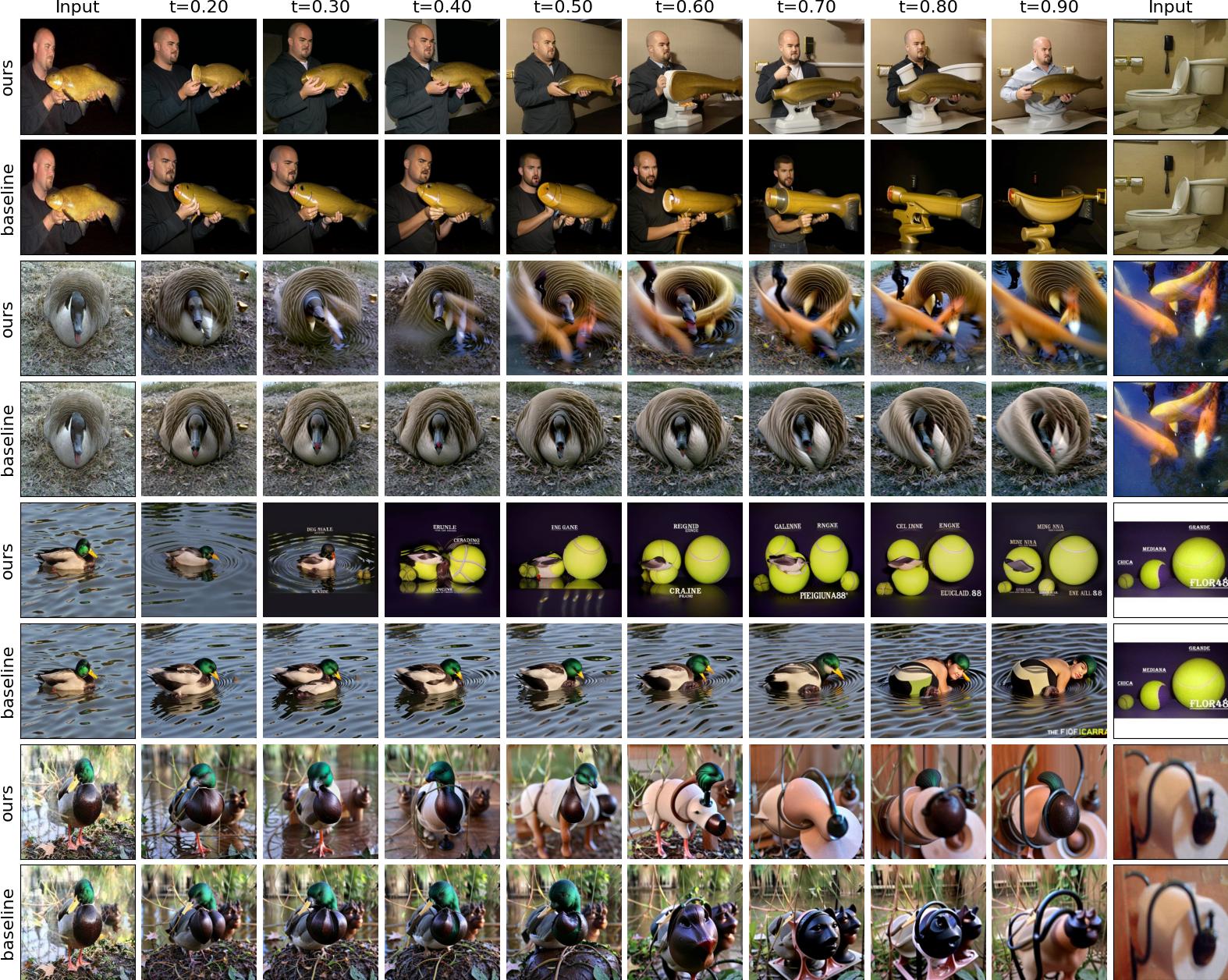}
    \caption{Interpolation results on Mood Space vs Original CLIP embedding space (baseline).}
    \label{fig:appendix_interpolation}
\end{figure*}

\clearpage
\begin{figure*}
    \centering
    \includegraphics[width=\linewidth]{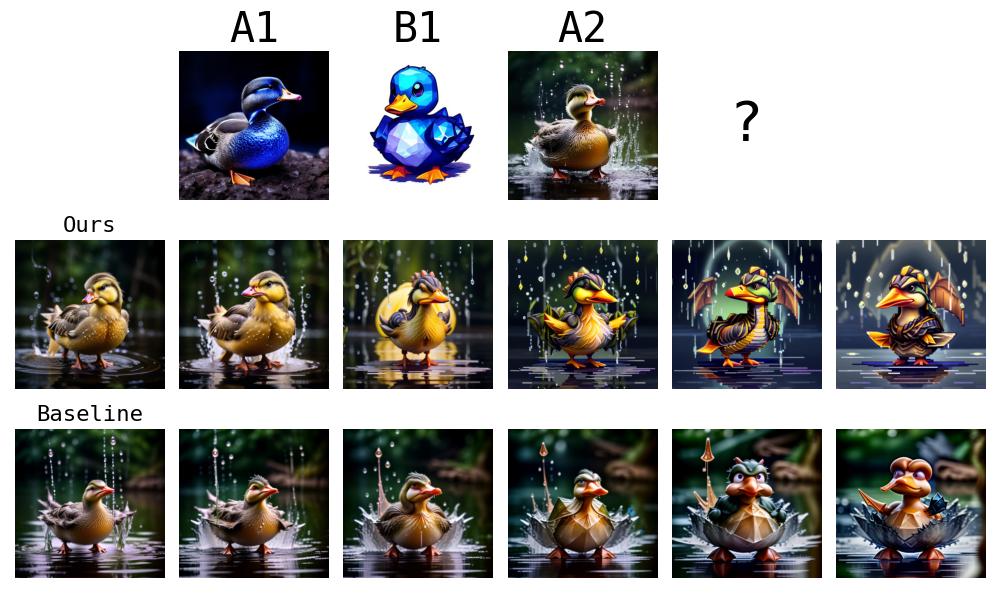}
    
    \vspace{1em} %
    
    \includegraphics[width=\linewidth]{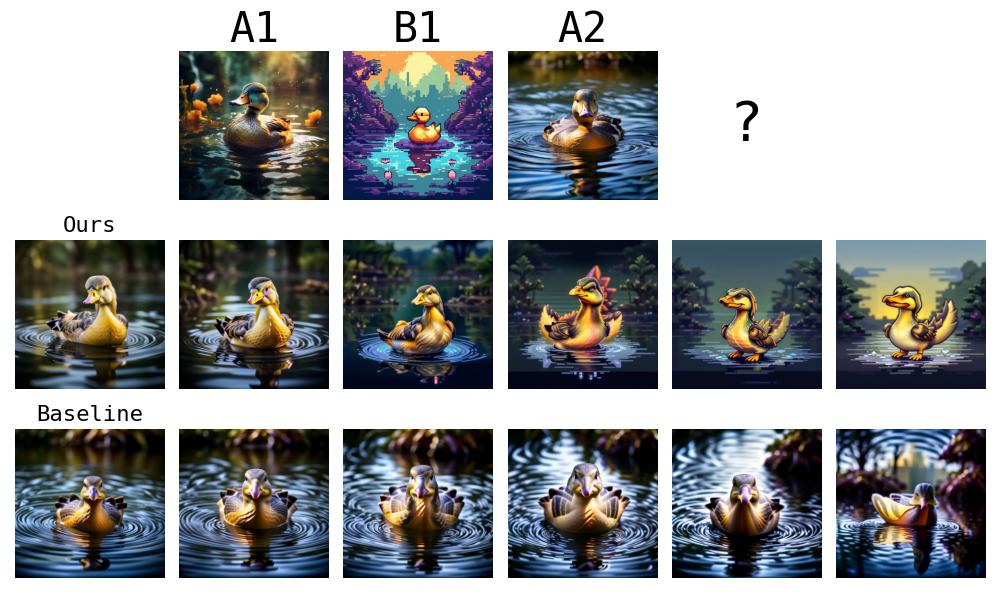}
    
    \caption{Analogy results on Mood Space vs Original CLIP embedding space (baseline). Top: First example. Bottom: Second example.}
    \label{fig:appendix_interpolation}
\end{figure*}

\clearpage
\begin{figure*}
    \centering
    \includegraphics[width=\linewidth]{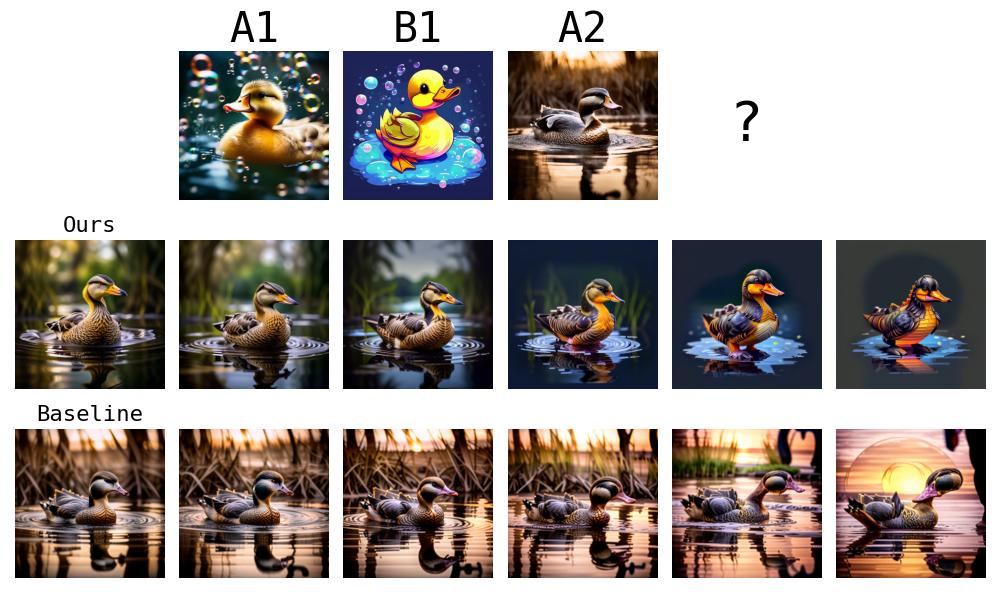}
    
    \vspace{1em} %
    
    \includegraphics[width=\linewidth]{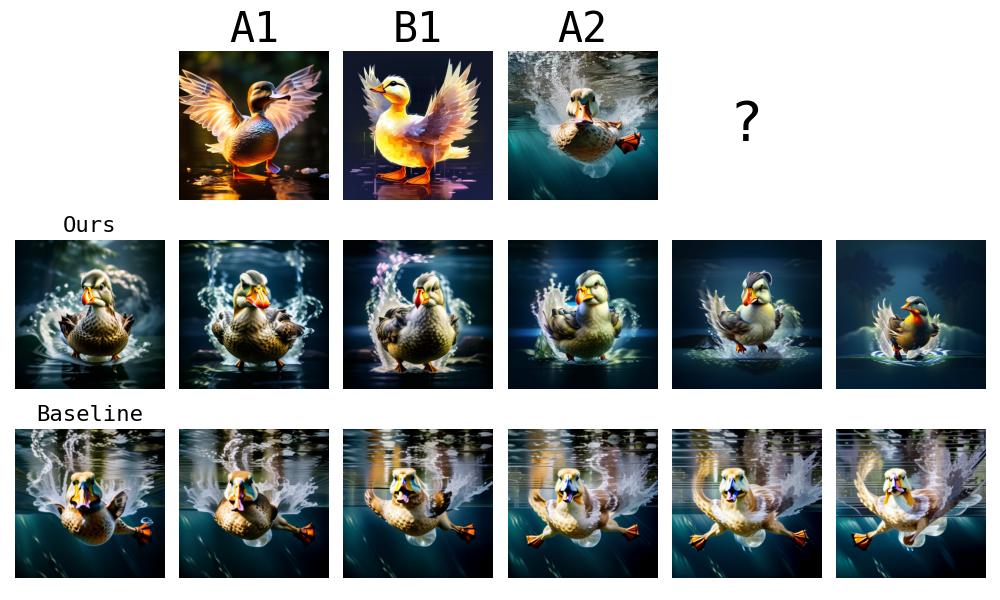}
    
    \caption{Analogy results on Mood Space vs Original CLIP embedding space (baseline). Top: First example. Bottom: Second example.}
    \label{fig:appendix_interpolation}
\end{figure*}

\clearpage
\begin{figure*}
    \centering
    \includegraphics[width=\linewidth]{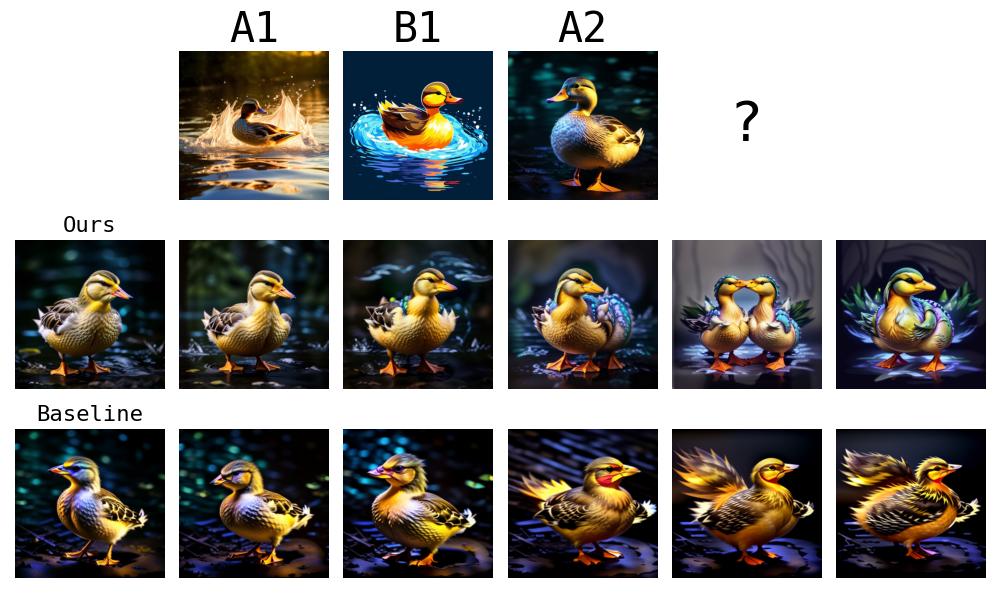}
    
    \vspace{1em} %
    
    \includegraphics[width=\linewidth]{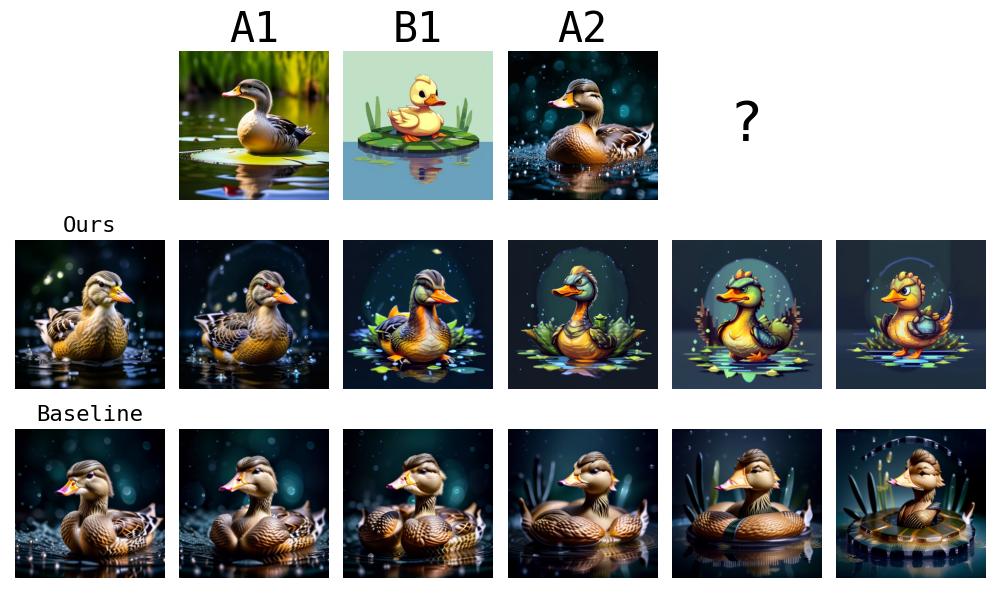}
    
    \caption{Analogy results on Mood Space vs Original CLIP embedding space (baseline). Top: First example. Bottom: Second example.}
    \label{fig:appendix_interpolation}
\end{figure*}

\clearpage
\begin{figure*}
    \centering
    \includegraphics[width=\linewidth]{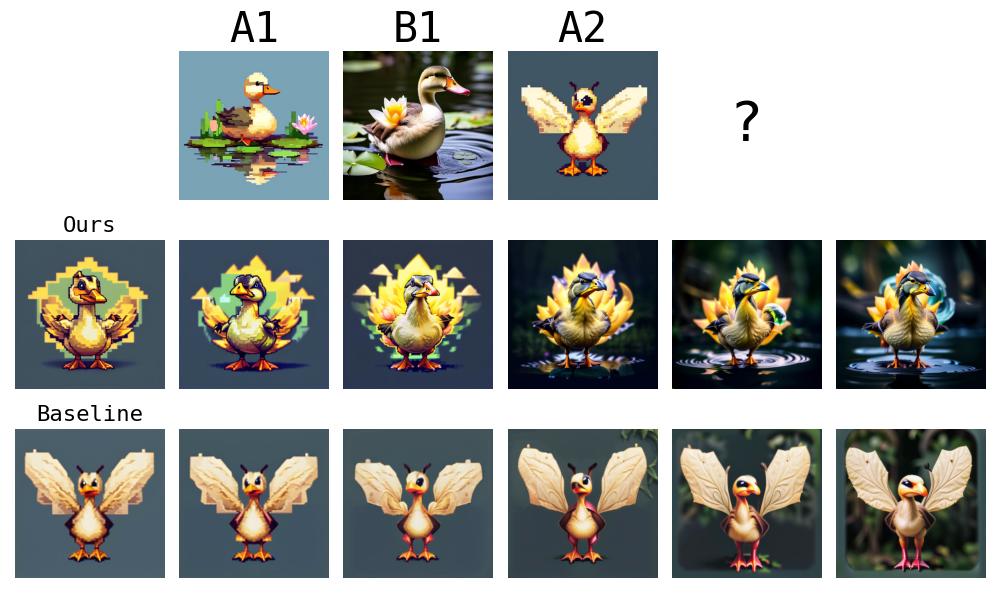}
    
    \vspace{1em} %
    
    \includegraphics[width=\linewidth]{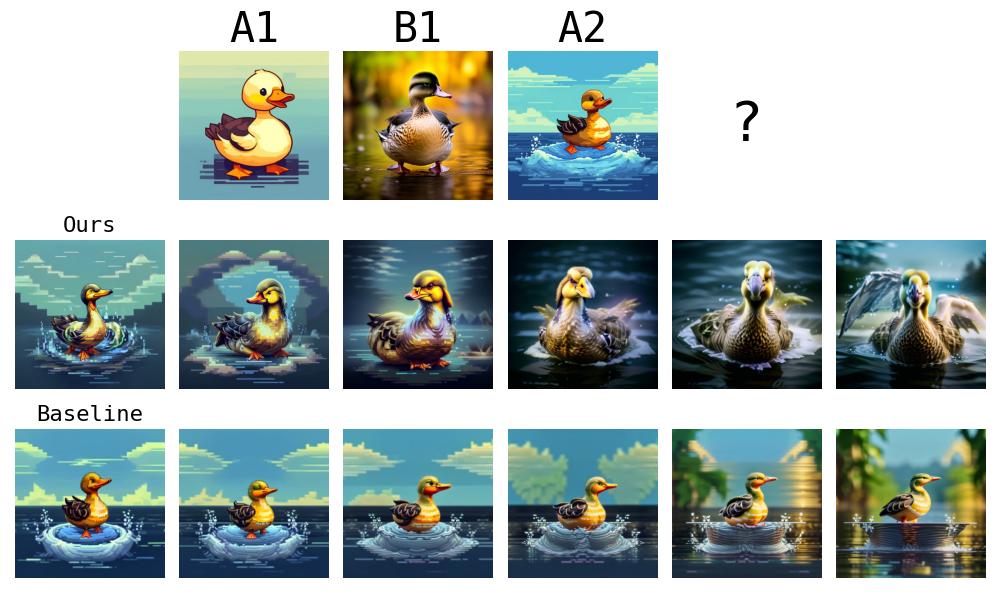}
    
    \caption{Analogy results on Mood Space vs Original CLIP embedding space (baseline). Top: First example. Bottom: Second example.}
    \label{fig:appendix_interpolation}
\end{figure*}

\clearpage
\begin{figure*}
    \centering
    \includegraphics[width=\linewidth]{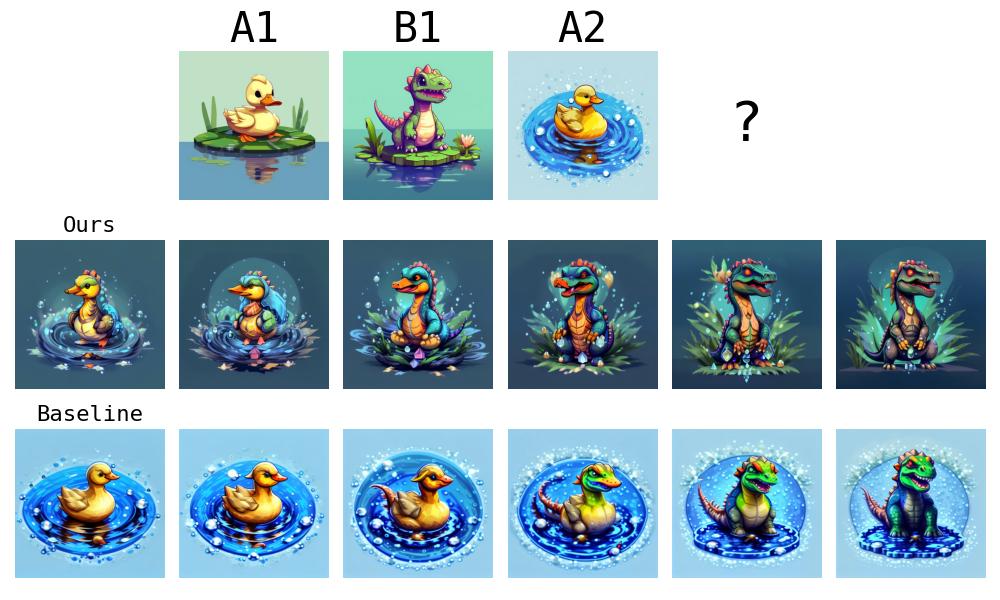}
    
    \vspace{1em} %
    
    \includegraphics[width=\linewidth]{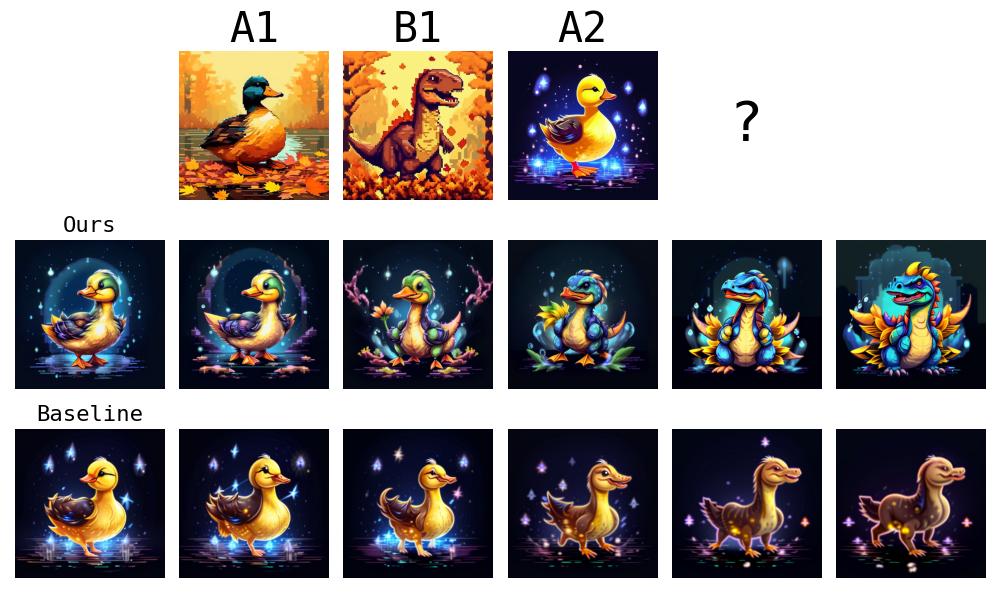}
    
    \caption{Analogy results on Mood Space vs Original CLIP embedding space (baseline). Top: First example. Bottom: Second example.}
    \label{fig:appendix_interpolation}
\end{figure*}

\end{appendix}

\end{document}